\documentclass[journal]{IEEEtran}

\ifCLASSOPTIONcompsoc
  \usepackage[nocompress]{cite}
\else
  \usepackage{cite}
\fi

\usepackage{times}
\usepackage{epsfig}
\usepackage{graphicx}
\usepackage{amsmath}
\usepackage{amssymb}
\usepackage{pdfpages}
\usepackage{caption}
\usepackage{subcaption}
\usepackage{xcolor}
\usepackage{hyperref}
\usepackage{amsfonts}
\usepackage{caption}
\usepackage{tabularx}
\usepackage{multirow}
\usepackage{gensymb}
\usepackage{bm}
\usepackage{color}

\ifCLASSINFOpdf
\else
\fi

\begin{document}
\title{Deep Learning for Inertial Positioning: A Survey}

\author{Changhao Chen and Xianfei Pan
\thanks{The authors are with the College of Intelligence Science and Technology, National University of Defense Technology, Changsha, 410073, China }
\thanks{Changhao Chen and Xianfei Pan are co-first authors. Changhao Chen is the corresponding author. (Email: changhao.chen66@outlook.com)}
\thanks{This work was supported by National Natural Science Foundation of China (NFSC) under the Grant Number of 62103427, 62073331, 62103430, 62103429. Changhao Chen is sponsored by the Young Elite Scientist Sponsorship Program by CAST (No. YESS20220181)}
}

\markboth{IEEE Transactions on Intelligent Transportation Systems, March~2024}%
{Shell \MakeLowercase{\textit{et al.}}: Bare Demo of IEEEtran.cls for IEEE Journals}



\maketitle

\begin{abstract}
Inertial sensors are widely utilized in smartphones, drones, vehicles, and wearable devices, playing a crucial role in enabling ubiquitous and reliable localization. Inertial sensor-based positioning is essential in various applications, including personal navigation, location-based security, and human-device interaction. However, low-cost MEMS inertial sensors' measurements are inevitably corrupted by various error sources, leading to unbounded drifts when integrated doubly in traditional inertial navigation algorithms, subjecting inertial positioning to the problem of error drifts. In recent years, with the rapid increase in sensor data and computational power, deep learning techniques have been developed, sparking significant research into addressing the problem of inertial positioning. Relevant literature in this field spans across mobile computing, robotics, and machine learning. In this article, we provide a comprehensive review of deep learning-based inertial positioning and its applications in tracking pedestrians, drones, vehicles, and robots. We connect efforts from different fields and discuss how deep learning can be applied to address issues such as sensor calibration, positioning error drift reduction, and multi-sensor fusion. This article aims to attract readers from various backgrounds, including researchers and practitioners interested in the potential of deep learning-based techniques to solve inertial positioning problems. Our review demonstrates the exciting possibilities that deep learning brings to the table and provides a roadmap for future research in this field.
\end{abstract}


%
\IEEEpeerreviewmaketitle

\section{Introduction}
\begin{figure}
    \centering
    \includegraphics[width=0.45\textwidth]{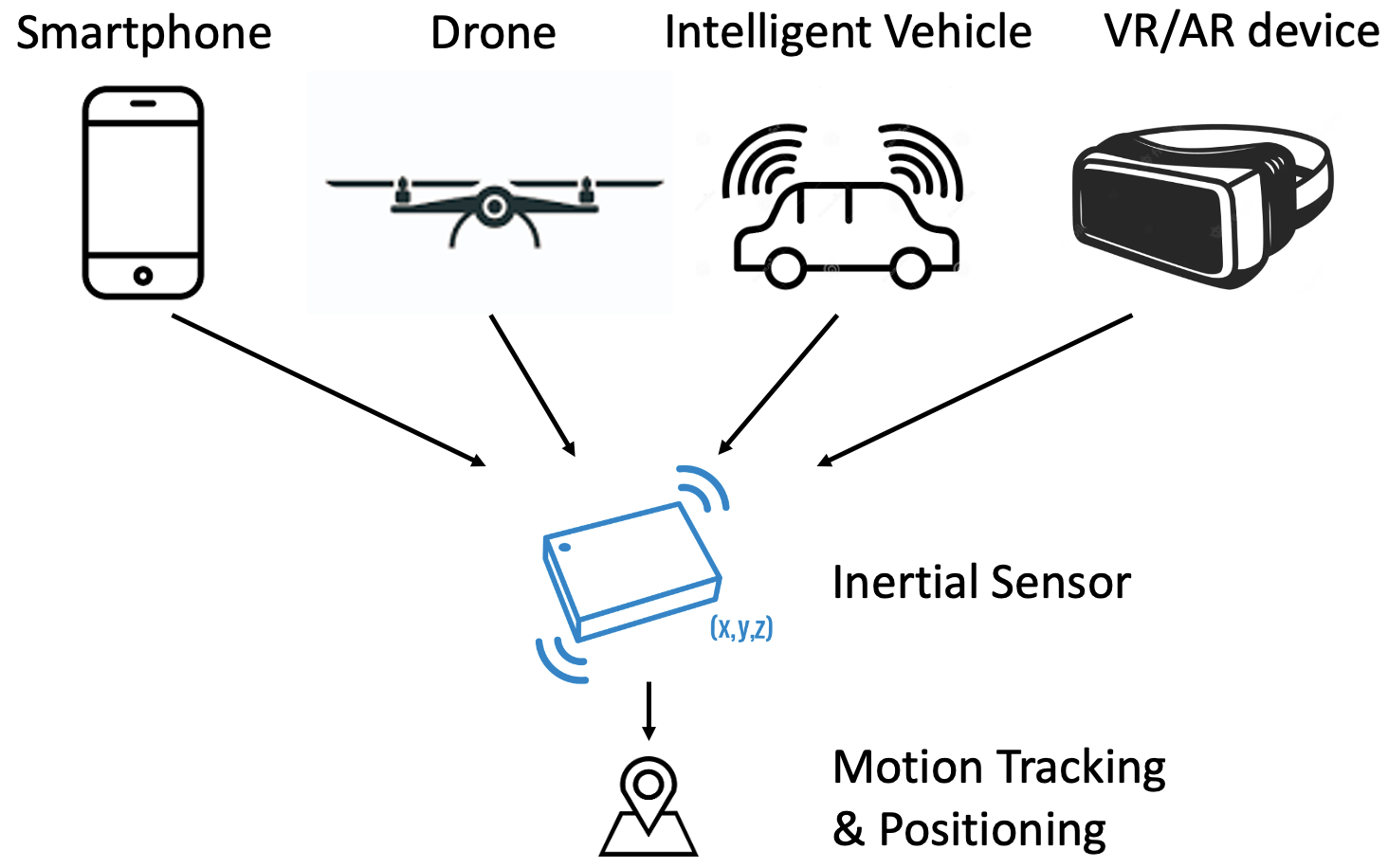}
    \caption{Inertial sensors are ubiquitous in modern platforms such as smartphones, drones, intelligent vehicles, and VR/AR devices. They play a critical role in enabling completely egocentric motion tracking and positioning, making them essential for a range of applications.}
    \label{fig: devices}
\end{figure}

\IEEEPARstart{T}{he} inertial Measurement Unit (IMU) is widely used in smartphones, drones, vehicles, and VR/AR devices. It continuously measures linear velocity and angular rate and tracks the motion of these platforms, as illustrated in Figure \ref{fig: devices}. With the advancements in Micro-Electro-Mechanical Systems (MEMS) technology, today's MEMS IMUs are small, energy-efficient, and cost-effective. Inertial positioning (navigation) calculates attitude, velocity, and position based on inertial measurements, making it a crucial element in various location-based applications, including locating and navigating individuals in transportation infrastructures (e.g., airports, train stations) \cite{puyol2014pedestrian}, supporting security and safety services (e.g., aiding first-responders) \cite{nilsson2014accurate}, enabling smart city/infrastructure, and facilitating human-device interaction \cite{bulling2014tutorial}. Compared to other positioning solutions such as vision or radio, inertial positioning is completely ego-centric, works indoors and outdoors, and is less affected by environmental factors such as complex lighting conditions and scene dynamics.

Unfortunately, the measurements obtained from low-cost MEMS IMUs are subject to several error sources such as bias error, temperature-dependent error, random sensor noise, and random-walk noise. In classical inertial navigation mechanisms, angular rates are integrated into orientation, and based on the acquired attitude, acceleration measurements are transformed into the navigation frame. Finally, the transformed accelerations are doubly integrated into locations \cite{Savage1998,Savage1998a}. Traditional inertial navigation algorithms are designed and described using concrete physical and mathematical rules. Under ideal conditions, sensor errors are small enough to allow hand-designed inertial navigation algorithms to produce accurate and reliable pose estimates. However, in real-world applications, inevitable measurement errors cause significant problems for inertial positioning systems without constraints, which can fail within seconds. In this process, even a minor error can be amplified exponentially, resulting in unbounded error drifts.

\begin{figure*}
    \centering
    \includegraphics[width=0.8\textwidth]{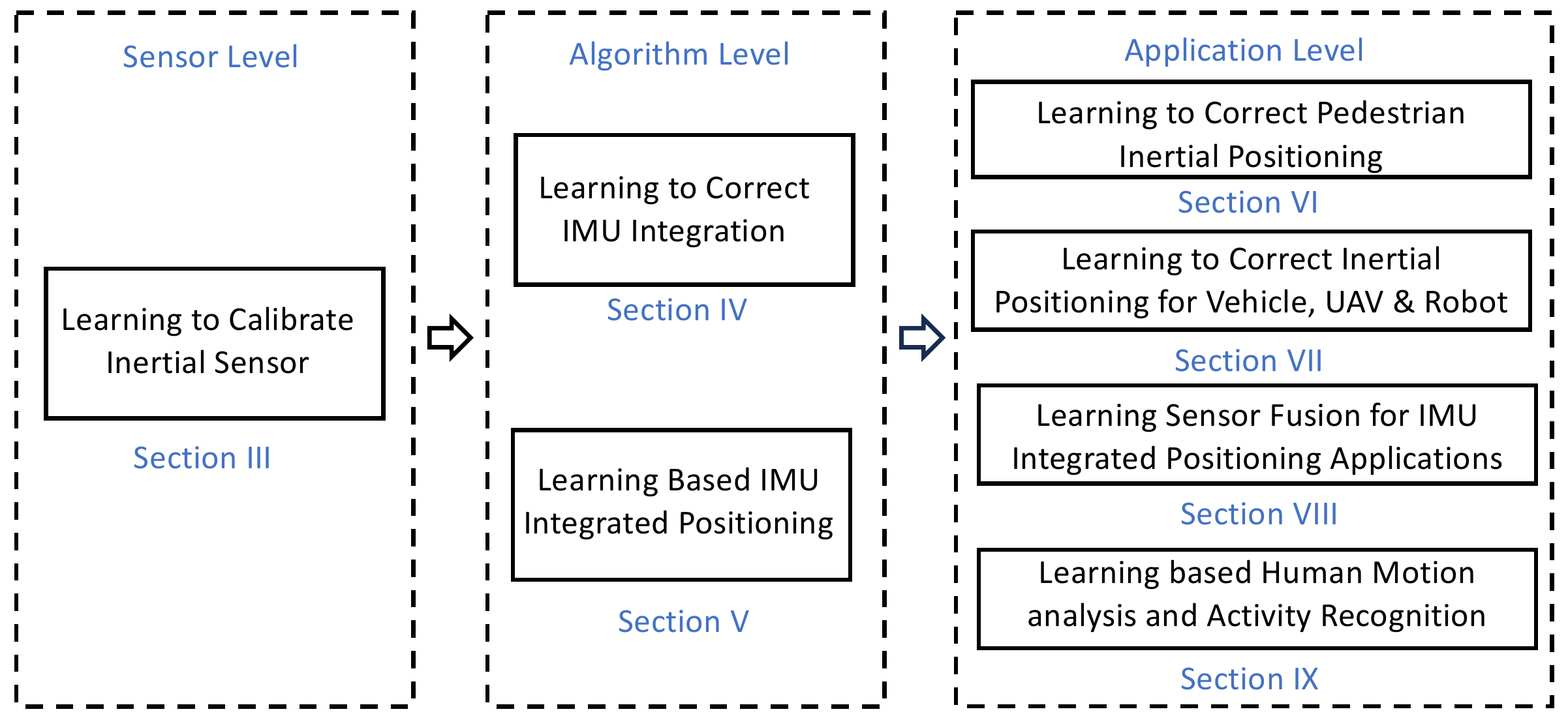}
    \caption{An overview of our survey structure.}
    \label{fig: overview}
\end{figure*}

Previous researchers have attempted to address the problem of error drifts in inertial navigation by incorporating domain-specific knowledge or other sensor. In the context of pedestrian tracking, exploiting the periodicity of human walking is important, and the process of pedestrian dead reckoning (PDR) involves detecting steps, estimating step length and heading, and updating the user's location to mitigate error drifts from exponential to linear increase \cite{harle2013survey}. Zero-velocity update (ZUPT) involves attaching the IMU to the user's foot and detecting the zero-velocity phase, which is then used in Kalman filtering to correct inertial navigation states \cite{Skog2010}. Platforms such as drones or robots equipped with other sensors such as cameras or LiDAR can significantly improve the performance of pure inertial solutions by effectively integrating inertial sensors with these modalities through filtering or smoothing \cite{Li2013b,Qin2018,xu2022fast}. However, these solutions have limitations in specific application domains and are unable to address the fundamental problem of inertial navigation.

Recently, deep learning has shown impressive performance in various fields, including computer vision, robotics, and signal processing \cite{bengio2017deep}. It has also been introduced to address the challenges of inertial positioning. Deep neural network models have been leveraged to calibrate inertial sensor noises, reduce the drifts of inertial navigation mechanisms, and fuse inertial data with other sensor information. These research works have attracted significant attention, as they show potential for exploiting massive data to generate data-driven models instead of relying on concrete physical or mathematical models. With the rapid development of deep learning techniques, learning-based inertial solutions have become even more promising.

\subsection{Taxonomy}
This survey aims to provide a comprehensive review of deep-learning-based approaches to inertial positioning, including measurement calibration, inertial positioning algorithms, and sensor fusion. 
To achieve this, we establish a taxonomy of existing deep learning-based inertial positioning approaches, and  conduct an analysis of their effectiveness at three levels: sensor level, algorithm level, and application level, as illustrated in Figure \ref{fig: overview}.

\begin{itemize}
    \item At the sensor level, deep learning is employed as a calibration method for inertial sensors. It effectively models error sources in inertial measurements and implicitly eliminates corrupted measurement errors and noise.
    \item At the algorithm level, deep neural networks are constructed to partially or completely replace modules in traditional inertial navigation mechanisms that enhance and correct IMU integration in general. Additionally, deep learning serves as a powerful tool for fusing inertial data with other sensor modalities such as cameras and LiDAR. Deep learning based IMU integration and fusion methods enable improved positioning accuracy and reliability.
    \item At the application level, we delve into specific use cases where deep learning methods can be applied to inertial positioning for pedestrians, vehicles, drones, and robots. For instance, we explore how learned motion patterns can enhance pedestrian dead reckoning (PDR) and zero-velocity update (ZUPT) algorithms.
\end{itemize}

Finally, we thoroughly discuss the advantages and limitations of existing works in this domain. We also identify the key challenges and future opportunities that lie ahead in this research direction.

\subsection{Comparison with Other Surveys}
 Compared with other deep learning surveys, such as those focused on object detection \cite{zhao2019object}, semantic segmentation \cite{hao2020brief}, and robotics \cite{sunderhauf2018limits}, \textit{survey on deep learning based inertial positioning is relatively scarce and hard to find}. While a broader survey on machine learning enhanced inertial sensing does exist \cite{li2021inertial}, our survey narrows the focus to deep learning based inertial positioning, providing deeper insights and analysis of the fast-evolving developments in this area over the past five years (2018-2022). Other relevant surveys, such as those focused on inertial pedestrian positioning \cite{harle2013survey}, indoor positioning \cite{farahsari2022survey}, step length estimation \cite{diez2018step}, and pedestrian dead reckoning \cite{wu2019survey}, do not cover recent deep learning based solutions. \textit{To the best of our knowledge, this article is the first survey that discusses deep learning based inertial positioning thoroughly and deeply.}

\subsection{Survey Organization}
The rest of this survey is organized as follows: Section II provides a brief overview of classical inertial navigation mechanisms. Sections III, IV, and V delve into methods and algorithms pertaining to sensor calibration based on deep learning, inertial navigation algorithms, and sensor fusion, respectively. Sections VI, VII, VIII, and IX explore specific applications of deep learning techniques within the realm of pedestrian inertial positioning, inertial positioning for vehicles, UAVs, and robots, IMU-integrated positioning, and human motion and activity recognition, respectively. Section X presents representative public datasets, evaluation metrics, and a performance comparison between learning-based and traditional inertial positioning models. Finally, Section XI concludes by discussing the benefits, challenges, and opportunities.

\section{Classical Inertial Navigation Mechanisms}
This section provides an overview of classical inertial navigation mechanisms and highlights their limitations. It begins by presenting the inertial measurement model and classical strapdown inertial navigation method. Subsequently, two solutions that aim to reduce the drifts of inertial navigation system, namely pedestrian dead reckoning (PDR) and zero-velocity update (ZUPT), are discussed, with a specific focus on their applicability in pedestrian tracking scenarios. The section finally introduces sensor fusion approaches that integrate inertial data with information from other sensors. 

\subsection{Inertial Measurement Model}
Inertial measurements acquired from low-cost MEMS IMUs are often corrupted by various types of error sources, resulting in unbounded error drifts when integrated in strapdown inertial navigation systems (SINS). These error sources can be classified into two categories: deterministic errors and random errors \cite{ru2022mems}. Deterministic errors comprise bias error, non-orthogonality error, misalignment error, scale-factor error, and temperature-dependent error. On the other hand, random errors include random sensor noise and random-walk noise resulting from long-term operation, which are challenging to model and eliminate.

Raw IMU measurements, i.e. accelerations $\hat{\mathbf{a}}$ and angular rates $\hat{\bm{\omega}}$, can be formulated by 
\begin{equation}
\label{eq: acc noise}
    \hat{\mathbf{a}} = \mathbf{a} +\mathbf{b}_{a} +\mathbf{n}_a
\end{equation}
\begin{equation}
\label{eq: gyro noise}
    \hat{\bm{\omega}} = \bm{\omega} +\mathbf{b}_{\omega} +\mathbf{n}_{\omega}
\end{equation}
where $\mathbf{b}_{a}$ and $\mathbf{b}_{\omega}$ are acceleration bias and gyroscope bias, $\mathbf{n}_a$ and $\mathbf{n}_{\omega}$ are additive noises above accelerometer and gyroscope.

Traditionally, it is important to calibrate inertial sensors before running an inertial navigation algorithm that involves integrating inertial data into system states. One effective tool for achieving this is the Allan variance method \cite{NaserEl-SheimyHaiyingHou2008}, which models the random process of inertial sensor errors.

\subsection{Strapdown Inertial Navigation System}
Inertial sensor measures linear accelerations $\mathbf{a}_b(t)$ and angular rates $\bm{\omega}_b^n(t)$ of attached user body at the timestep $t$. $b$ represents the body frame, while $n$ denotes the navigation (world) frame, i.e. the navigation frame. $\bm{\omega}_b^n(t)$ means that the angular rates of body frame with respect to the navigation frame.
To simplify inertial motion model, this article assumes that the biases and noises of sensor in Equation \ref{eq: acc noise} and \ref{eq: gyro noise} have been removed in the stage of inertial sensor calibration. $(\mathbf{R}, \mathbf{p})$ are defined orientation and position variables.
From the kinematic model of IMU, we can have
\begin{equation}
    \label{eq: kinematic1}
    \begin{cases}
    \mathbf{R}_b^n(t+1) = \mathbf{R}_b^n (t)  \mathbf{R}_{b_{t+1}}^{b_{t}} \\
    \mathbf{v}_n(t+1) = \mathbf{v}_n(t) + \mathbf{a}_n(t) dt \\
    \mathbf{p}_n(t+1) = \mathbf{p}_n(t) + \mathbf{v}_n(t) dt + \frac{1}{2} \mathbf{a}_n(t) dt^2
    \end{cases}
\end{equation}
where  $\mathbf{a}_n$, $\mathbf{v}_n$, $\mathbf{p}_n$ are acceleration, velocity and position in the navigation frame, $\mathbf{R}_b^n$ represents the rotation from the body frame to the navigation frame.

Firstly, orientation is updated by inferring the rotation matrix $ \bm{\Omega(t)}$ via Rodriguez formula:
\begin{equation}
\label{eq: update}
    \begin{split}
        \bm{\Omega}(t) &= \mathbf{R}_{b_{t+1}}^{b_{t}} \\
        &= \mathbf{I} +\sin(\bm{\sigma}) \frac{[\bm{\sigma} \times]}{\bm{\sigma}} + (1-\cos(\bm{\sigma}))\frac{[\bm{\sigma} \times]^2}{\bm{\sigma}^2},
    \end{split}
\end{equation}
where rotation vector $\bm{\sigma} = \bm{\omega}(t) dt$.

To update velocity,
the accelerations in navigation frame can be expressed as a function of measured accelerations, i.e.
\begin{equation}
    \mathbf{a}_n(t) = \mathbf{R}_b^n(t-1)\mathbf{a}_b(t) - \mathbf{g}_n
\end{equation}

Then, the accelerations in navigation frame $\mathbf{a}_n(t)$ are integrated into the velocity in the navigation frame $\mathbf{v}_n(t)$, and the location $\mathbf{p}_n(t)$ is finally updated by integrating the velocity via Equation \ref{eq: update}. 

As we can see, in this process, even a small measurement error can be exponentially amplified, leading to the problem of inertial error drifts. In the past, high-precision inertial sensors such as laser or fiber inertial sensors could keep the measurement error small enough to maintain the accuracy of INS. However, due to the size and cost limitations of current MEMS IMUs, compensation methods are necessary to mitigate the corresponding error drifts. One approach is to introduce domain-specific knowledge or other sensor information.

\subsection{Domain Specific Knowledge}
\subsubsection{Pedestrian Dead Reckoning}
Pedestrian dead reckoning (PDR) is a method that leverages domain-specific knowledge about human walking to track pedestrian motion. PDR comprises three main steps: step detection, heading and stride length estimation, and location update \cite{harle2013survey}. In step detection, PDR uses the threshold of inertial data to identify step peaks or stances and segment the corresponding inertial data. Dynamic stride length estimation is then achieved via an empirical formula, known as the Weinberg formula \cite{weinberg2002using}, which considers the segmented accelerations and user's height. Similar to SINS, heading estimation is done by integrating gyroscope signals into orientation changes and adding orientation changes to the initial orientation to obtain the current heading. Finally, the estimated heading and stride length are used to update the pedestrian's location. By avoiding double integration of accelerations and incorporating a reliable stride estimation model, PDR effectively reduces inertial positioning drifts. However, inaccurate step detection and stride estimation can still occur, leading to large system error drifts. Moreover, PDR is limited to pedestrian navigation as it depends on the periodicity of human walking.

\subsubsection{Zero-velocity Update}
The Zero-velocity update (ZUPT) algorithm is designed to compensate for the errors of SINS by identifying the still phase of human walking and using zero-velocity as observations in a Kalman filter \cite{Skog2010}. To facilitate the detection of the still phase, the IMU is typically attached to the user's foot, as it undergoes significant motion and reflects walking patterns well. Techniques such as peak-detection \cite{fang2005design}, zero-crossings \cite{goyal2011strap}, or auto-correlation \cite{huang2016exploiting} can be used to analyze the inertial data and segment the zero-velocity phase. Once the still phase is detected, zero-velocity is used as pseudo-measurements in the filtering process, thereby limiting the error drifts of open-loop integration. However, the effectiveness of ZUPT depends on the assumption that the user's foot remains completely still, and any incorrect still phase detection or small motion disturbances can cause navigation system drifts. Additionally, ZUPT is limited to pedestrian tracking.

\subsection{Integrating IMU with Other Sensors}
Integrating the IMU with other sensors, such as camera \cite{Qin2018}, LiDAR \cite{xu2022fast}, UWB \cite{feng2020kalman}, and magnetometer \cite{yang2021robust}, can provide promising results as it allows for exploiting their complementary properties. By fusing the data from multiple sensors, the accuracy and robustness of pose estimation can be significantly improved, making it a general solution for all platforms. However, in some scenarios, certain sensors, such as visual perception, may not be available or highly dependent on the environment, which can negatively affect the egocentric property of inertial positioning. Additionally, in sensor fusion approaches, it is essential to consider various factors such as sensor calibration, initialization, and time-synchronization.

\begin{table*}[t]
\caption{A summary of existing methods on deep learning based inertial sensor calibration.}
\label{tb: inertial}
    \small
    \begin{center}
\begin{tabular}{c c c c c c}
 \hline
name  & year & sensor & model & learning &  target \\
                               \hline
Xiyuan et al.\cite{xiyuan2003modeling}   & 2003 & gyro & 1-layer NN & SL & gyro drifts compensation\\
Chen et al. \cite{chen2018improving} & 2018 & gyro, acc & ConvNet & SL  & inertial noise compensation\\
Esfahani et al. \cite{esfahani2019orinet} & 2019 & gyro & LSTM & SL & gyroscope calibration \\
Nobre et al. \cite{nobre2019learning} & 2019 & gyro, acc & Deep Q-Network &  RL &  optimal calibration parameters  \\
Brossard et al. \cite{brossard2020denoising} & 2020 & gyro &  ConvNet & SL & gyro corrections \\
Zhao et al.\cite{zhao2020learning} & 2020 & gyro & LSTM & SL & gyroscope calibration\\
Huang et al.\cite{huang2022mems}& 2022 & gyro & Temporal ConvNet & SL & gyroscope calibration \\
Calib-Net \cite{li2022calib} & 2022 & gyro & Dilated ConvNet & SL & gyroscope denoising\\
\hline
\end{tabular}
    \begin{itemize}
             \footnotesize{
               \item \textit{Years} indicates the publication year of each work.
                \item \textit{Sensors} indicates the sensors involved in each work. gyro and acc represent gyroscope and accelerometer respectively.
                \item \textit{Model} indicates which module the framework consists of.
                \item \textit{Learning} indicates how to train neural networks. SL and RL represent Supervised Learning and Reinforcement Learning.
                \item \textit{Target} indicates what the model aims to solve or produce.
           }
   \end{itemize}
\end{center}
\end{table*}

\subsection{Discussion}
As previously mentioned, classical inertial navigation methods are designed to solve specific problems within their respective domains. However, their performance is often limited due to real-world issues such as imperfect modeling, measurement errors, and environmental influences, resulting in inevitable error drifts. Researchers in the field of inertial navigation are therefore constantly searching for ways to build models that can tolerate measurement errors and mitigate system drifts. In addition to relying on Newtonian physical rules, it has been observed that domain-specific knowledge, whether it be an experienced human walking model or scene geometry, can serve as a useful constraint in reducing the error drifts of inertial positioning systems. One potential approach to improving inertial positioning accuracy and robustness is to exploit massive inertial data to extract domain-specific knowledge and construct a data-driven model. In the next sections, we will delve deeper into this problem and explore potential solutions.

\begin{figure}
    \centering
    \includegraphics[width=0.5\textwidth]{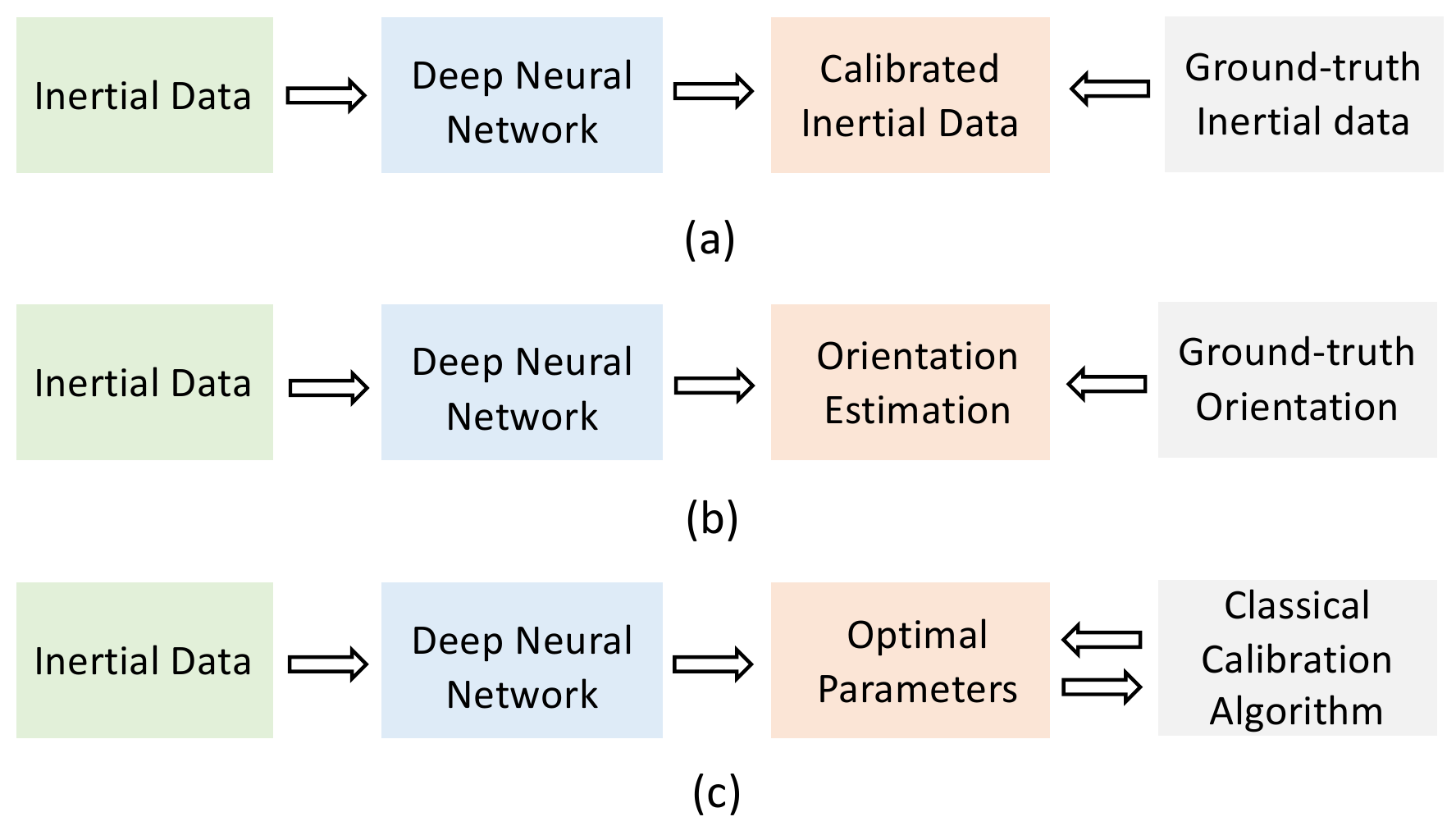}
    \caption{An overview of existing deep learning based inertial sensor calibration methods}
    \label{fig: calibration}
\end{figure}

\section{Learning to Calibrate Inertial Sensor}
Inertial measurements obtained from low-cost IMUs are often affected by various sources of noise, making it challenging to distinguish the true values from the sources of error. The error sources are a complex interplay of deterministic and random factors, further complicating the issue. To address the impact of measurement errors, the powerful nonlinear approximator capabilities of deep neural networks can be exploited. A natural approach is to develop a deep neural network that receives the raw inertial measurements as input and produces the calibrated inertial measurements as output, representing the actual platform motion. By training this neural model on labeled datasets using stochastic gradient descent (SGD) \cite{amari1993backpropagation}, the inertial measurement errors can be implicitly learned and corrected by the neural network. It is important to note that the quality of the collected training dataset has a significant impact on the performance of the model.

Before the age of deep learning, attempts were made to use neural networks to learn the measurement errors of inertial sensors. For example, a 1-layer artificial neural network (ANN) \cite{jain1996artificial} is proposed to model the distribution of gyro drifts, and is able to successfully approximate gyro drifts with such a 'shallow' network \cite{xiyuan2003modeling} . This method has an advantage over Kalman filtering (KF) based calibration methods in that it does not require setting hyper-parameters before use, such as the sensor noise matrix in KF.

 \begin{figure}
    \centering
\includegraphics[width=0.4\textwidth]{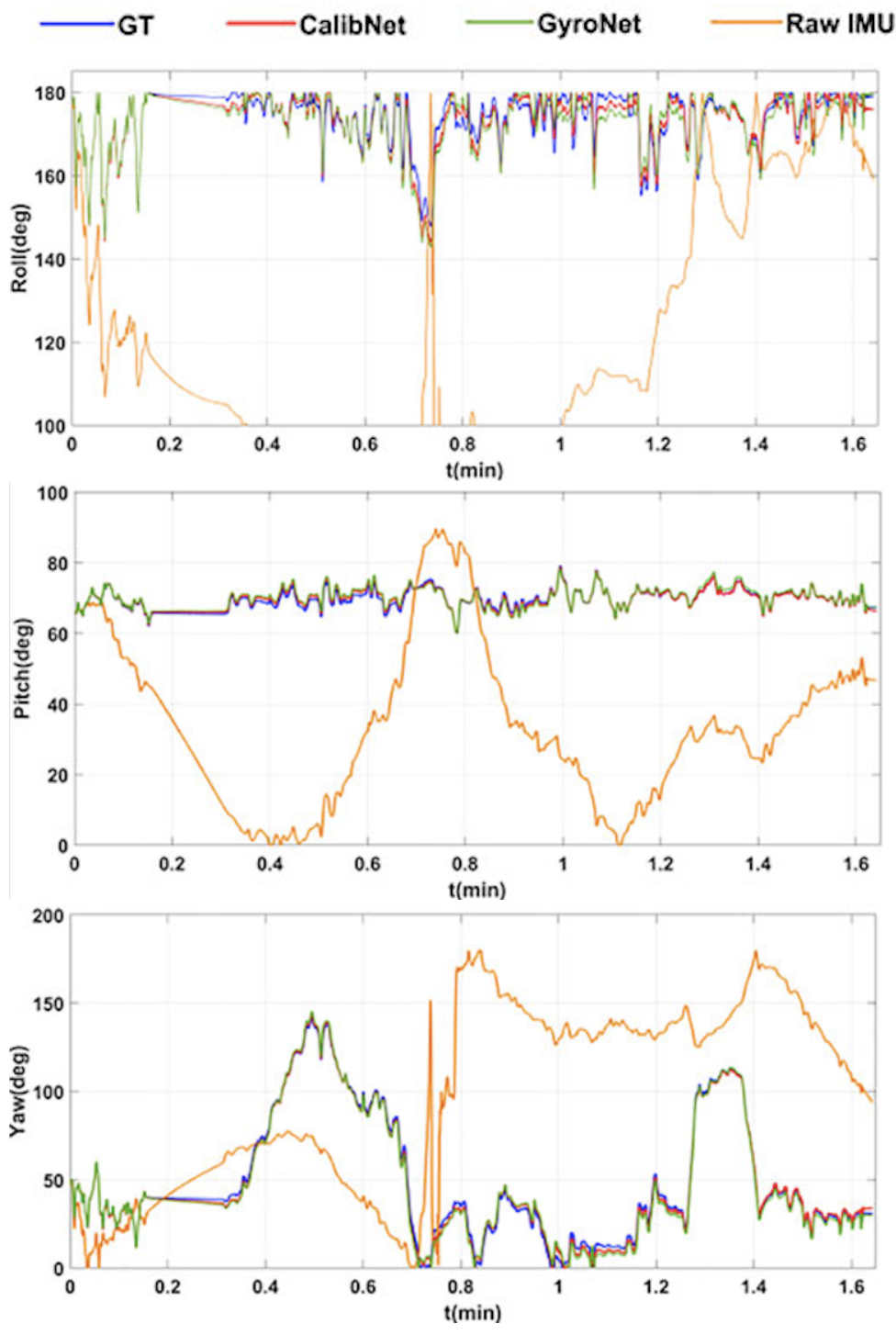}
    \caption{An example of gyro calibration results (reprint from Calib-Net \cite{li2022calib}). Compared with raw IMU integration, deep learning based calibration models significantly reduce attitude drifts.}
    \label{fig: calibration result}
\end{figure}

\begin{table*}[t]
    \caption{A summary of existing methods on deep learning based inertial positioning.}
    \label{tb: positioning}
    \small
    \begin{center}
    \begin{tabular}{c c c c c c}
 \hline
name  & Year & Carrier & model & learning & target \\
\hline
IONet \cite{chen2018ionet} & 2018   & Pedestrian, Trolley  & LSTM & SL & location displacement \\
RIDI \cite{yan2018ridi} & 2018 & Pedestrian & SVM, SVR & SL & velocity for inertial data calibration \\
Cortes et al.\cite{cortes2018deep} & 2018 & Pedestrian & ConvNet & SL & velocity to constrain system drifts\\
Wagstaff et al.\cite{wagstaff2018lstm} & 2018 & Pedestrian & LSTM  & SL &  zero-velocity detection for ZUPT\\
Chen et al.\cite{chen2019motiontransformer} & 2019 & Pedestrian, Trolley & LSTM & TL & location displacement \\
AbolDeepIO \cite{esfahani2019aboldeepio} & 2019 & UAV & LSTM & SL & location displacement\\
RINS-W \cite{brossard2019rins} & 2019 & Vehicle & RNN & SL &  zero-velocity dection for KF\\
Feigl et al. \cite{feigl2019bidirectional} & 2019 & Pedestrian & LSTM & SL & walking velocity\\
Wang et al. \cite{wang2019pedestrian} & 2019 & Pedestrian & LSTM & SL & walking heading for ZUPT\\
Yu et al. \cite{yu2019azupt} & 2019 & Pedestrian & ConvNet  & SL & adaptive zero-velocity detection \\
TLIO \cite{liu2020tlio} & 2020 & Pedestrian & ConvNet  & SL & 3D displacement and uncertainty for EKF\\
LIONet \cite{chen2020deep} & 2020 & Pedestrian & Dilated ConvNet  & SL & lightweight inertial model\\
RoNIN\cite{herath2020ronin} & 2020 & Pedestrian & LSTM, TCN & SL & velocity for inertial data calibration \\
Brossard et al.\cite{brossard2020ai} & 2020 & Vehicle & ConvNet & SL & co-variance noise for KF\\
StepNet\cite{klein2020stepnet} & 2020 & Pedestrian & ConvNet, LSTM & SL & dynamic step length for PDR\\
Wang et al.\cite{wang2020pedestrian} & 2020 & Pedestrian & ConvNet & SL & measurement noise for Kalman Filter\\
ARPDR \cite{teng2020arpdr} & 2020 & Pedestrian & TCN & SL & stride length and walking heading for PDR\\
IDOL\cite{sun2021idol} & 2021 & Pedestrian & LSTM  & SL & device orientation and location\\
PDRNet\cite{asraf2021pdrnet} & 2021 & Pedestrian & ConvNet & SL & step length and heading for PDR\\
Buchanan et al. \cite{buchanan2022learning} & 2021 & Legged Robot & ConvNet & SL & integrate location displacement with leg odometry \\
Zhang et al.\cite{zhang2021imu} & 2021 & Vehicle, UAV & RNN & SL & independent motion terms 
\\
Gong et al. \cite{gong2021robust} & 2021 & Pedestrian & LSTM & SL & fusing inertial data from two devices\\
NILoc\cite{herath2022neural} & 2022 & Pedestrian & ConvNet & SL & inertial relocalization\\
RIO\cite{cao2022rio} & 2022 & Pedestrian & DNN  & UL & rotation-equivariance as supervision signal\\
Wang et al. \cite{wang2022magnetic} & 2022 & Pedestrian & DNN & SL & efficient and low-latent model \\
TinyOdom\cite{saha2022tinyodom} & 2022 & Pedestrian, Vehicle & TCN+NAS & SL & deployment on resource-constrained device\\
CTIN\cite{rao2022ctin} & 2022 & Pedestrian & Transformer & SL & velocity and trajectory prediction \\
DeepVIP\cite{zhou2022deepvip} & 2022 & Vehicle & ConvNet, LSTM & SL & velocity and heading for car localization\\
Bo et al.\cite{bo2022mode} & 2022 & Pedestrian & ConvNet & TL & model-independent stride learning\\
OdoNet\cite{tang2022odonet} & 2022 & Vehicle & ConvNet & SL & speed learning for ZUPT\\
A2DIO\cite{wang2022a2dio} & 2022 & Pedestrian & ConvNet, LSTM & SL & pose invariant odometry\\
LLIO \cite{wang2022llio} & 2022 & Pedestrian & MLP & SL & 3D displacement for lightweight odometry \\
Liu et al. \cite{liu2023smartphone} & 2023 & Pedestrian & TCN & SL & general model trained on large dataset\\
\hline
\end{tabular}
\end{center}
\begin{itemize}
    \footnotesize{
    \item \textit{Year} indicates the publication year of each work.
        \item \textit{Carrier} indicates the platform running inertial navigation.  
        \item \textit{Model} indicates which module the framework consists of.
        \item \textit{Learning} indicates how to train neural networks. SL, TL and UL represent Supervised Learning, Transfer Learning and Unsupervised Learning.
        \item \textit{Target} indicates what the model aims to solve or produce.
    }
    \end{itemize}
\end{table*}

In recent years, there has been increasing interest in using deep neural networks (DNN) with multiple layers to solve the inertial sensor calibration problem. With the addition of more layers, neural networks become more expressive and can learn complex relationships between the raw inertial measurements and the true motion of the vehicle. One approach, proposed by \cite{chen2018improving}, uses a Convolutional Neural Network (ConvNet) to remove error noises from inertial measurements. They collected inertial data from two grades of IMU under given constant accelerations and angular rates. The ConvNet framework takes raw inertial measurements (from low-precision IMU) as inputs and tries to output acceleration and angular rate references (from high-precision IMU). Their experiment shows that deep learning can remove some of the sensor error and improve test accuracy. However, this work has not been validated in a real navigation setup, and thus it cannot demonstrate how learning-based sensor calibration reduces error drifts in inertial navigation. Both of the mentioned methods require reference data from high-precision IMUs as labels to train the networks, as shown in Figure \ref{fig: calibration} (a). However, acquiring reference data from high-precision IMUs can be costly.

 In addition to directly learning from pseudo ground-truth IMU labels, another approach is to enable neural network-based calibration models to produce inertial data that can be integrated into more accurate orientation estimation. This is illustrated in Figure \ref{fig: calibration} (b). By producing more accurate orientation values, the neural network implicitly removes the corrupted noises above inertial data. For example, OriNet \cite{esfahani2019orinet} inputs 3-dimensional gyroscope signals into an LSTM network \cite{hochreiter1997long} to obtain calibrated gyroscope signals, which are then integrated with the orientation at the previous timestep to generate orientation estimates at the current timestep. A loss function between orientation estimates and real orientation is defined and minimized for model training. OriNet has been evaluated on a public drone dataset, demonstrating an improvement in orientation performance of approximately 80\%. A similar approach is \cite{brossard2020denoising}, who calibrates gyroscope using ConvNet, reporting good attitude estimation accuracy.
Calib-Net \cite{li2022calib} is another ConvNet framework that denoises gyroscope data by extracting effective spatio-temporal features from inertial data. Calib-Net is based on dilation ConvNet \cite{yu2016multi} to compensate the gyro noise, as illustrated in Figure \ref{fig: calibration result}. This model is able to significantly reduce orientation error compared to raw IMU integration. When this learned inertial calibration model is incorporated into a visual-inertial odometry (VIO), it further improves localization performance and outperforms representative VIOs such as VINS-mono \cite{Qin2018}. Other efforts in this direction include works by \cite{zhao2020learning,huang2022mems}.

Instead of directly calibrating inertial sensors with DNNs, some researchers have explored using DNNs to generate parameters that improve classical calibration algorithms, as shown in Figure \ref{fig: calibration} (c). One example is the work by \cite{nobre2019learning}, who models inertial sensor calibration as a Markov Decision Process and proposes to use deep reinforcement learning \cite{sutton2018reinforcement} to learn the optimal calibration parameters. The authors demonstrated the effectiveness of their approach in calibrating inertial sensors for a visual-inertial odometry (VIO) system.

\begin{figure*}
    \centering
    \includegraphics[width=0.8\textwidth]{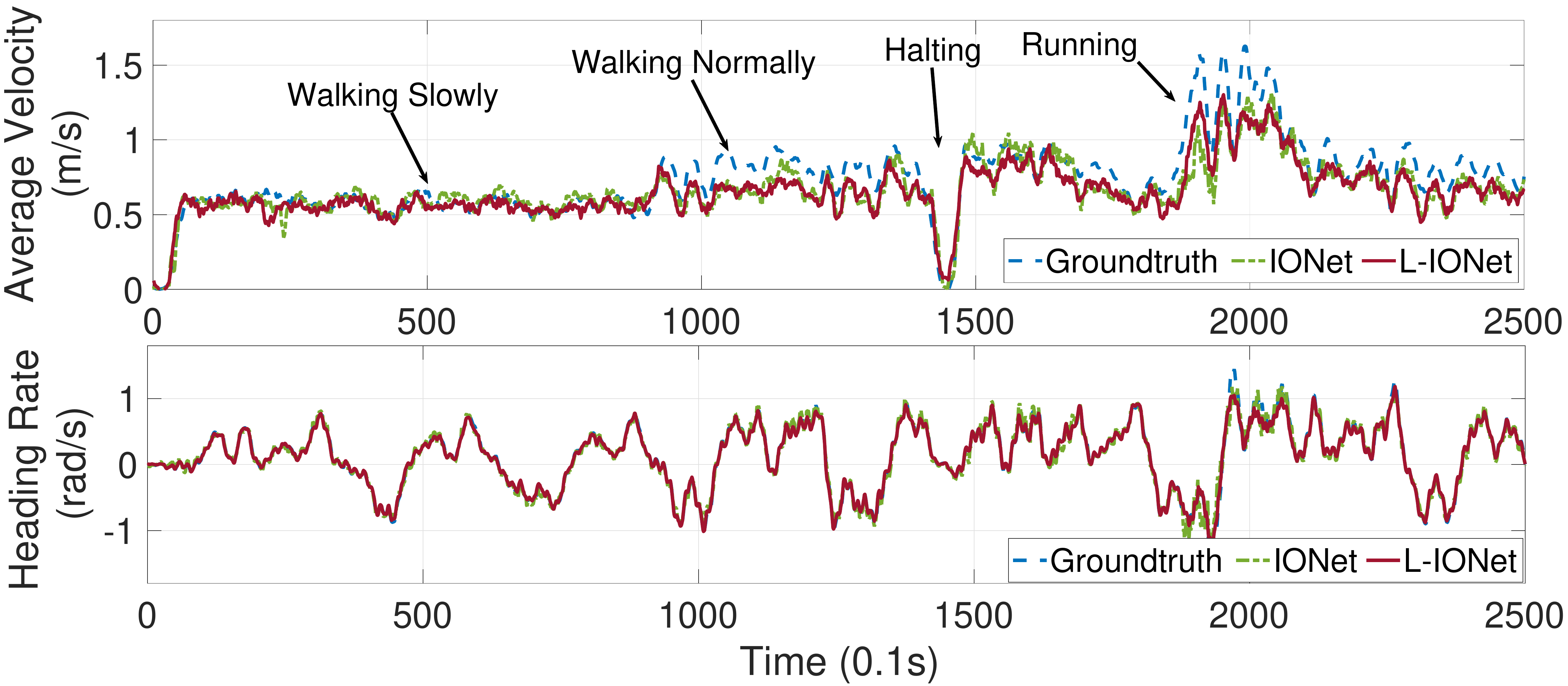}
    \caption{The velocity of attached platform can be inferred from a sequence of inertial measurements via deep neural networks. (reprint from L-IONet \cite{chen2020deep})}
    \label{fig: velocity}
\end{figure*}

\begin{figure}
    \centering
    \includegraphics[width=0.5\textwidth]{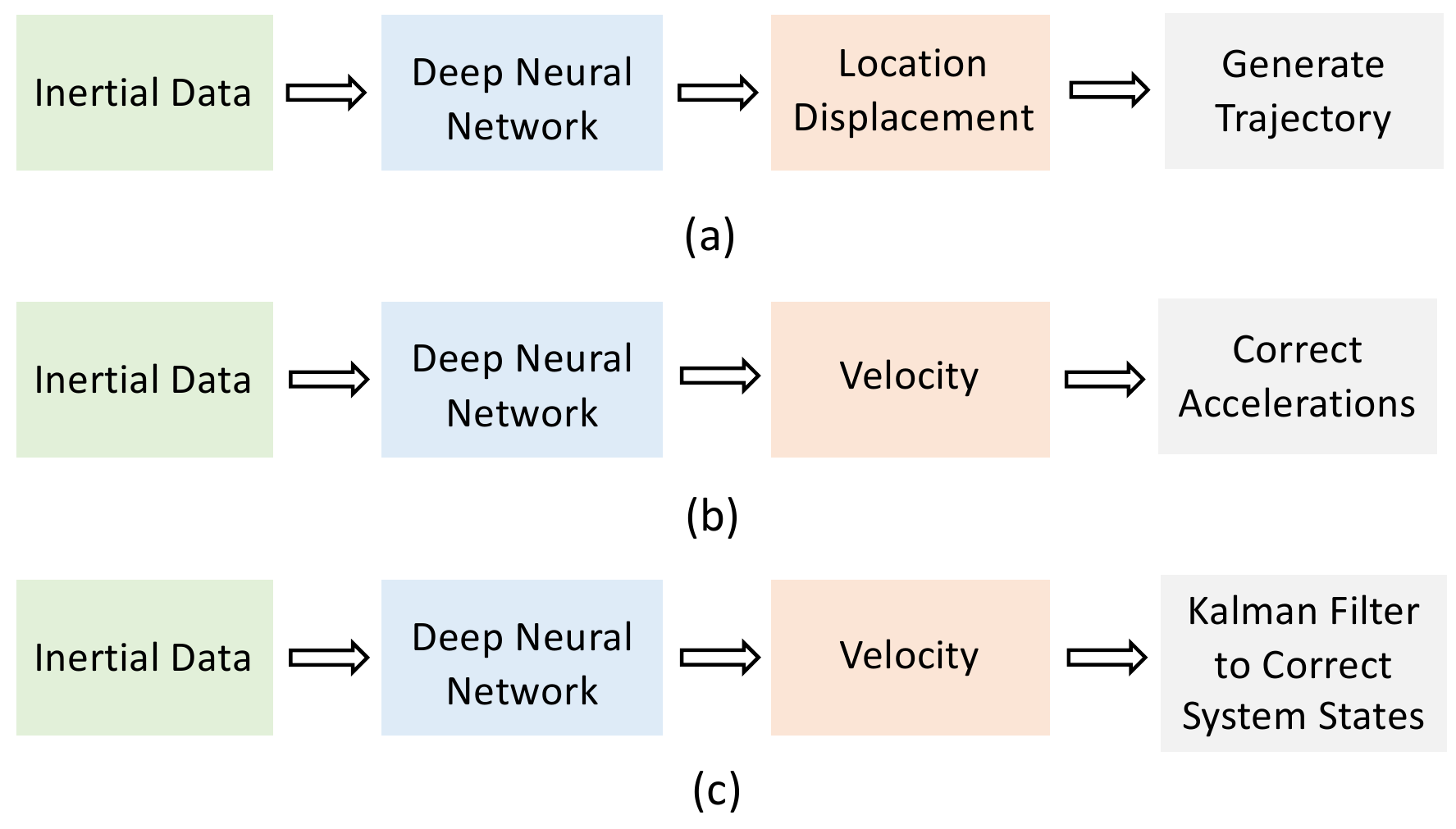}
    \caption{An overview of existing methods on learning to correct IMU integration}
    \label{fig: integration}
\end{figure}

As discussed above, deep learning-aided inertial sensor calibration methods (listed in Table \ref{tb: inertial}) have shown promising results in removing corrupted sensor noises and improving the accuracy of inertial positioning systems. These methods do not require human intervention and can automatically learn error models. However, it is important to note that the learned error model is typically dependent on the specific sensor or platform used. Therefore, a change in sensor or user can result in different data distributions, leading to reduced performance of the learned model. Additionally, further analysis is needed to determine which types of noise can be effectively removed by learning-based calibration methods.

\section{Learning to correct IMU Integration}
\label{sec: ins}
In addition to sensor calibration, researchers are exploring various methods for using deep learning to construct inertial positioning models that can either partially or completely replace classical inertial navigation mechanisms. This section provides an overview of how deep learning can be used to correct IMU integration in general. Next sections will discuss deep learning approaches for pedestrian tracking applications, and  present deep inertial solutions for vehicles, UAVs, and robots. A summary of existing works and their contributions is provided in Table \ref{tb: positioning}.


 In deep learning-based inertial positioning approaches, a user's absolute velocity can be inferred from a sequence of IMU data using a deep neural network. This velocity information can then be used as a key constraint to reduce the drifts in IMU double integration. Figure \ref{fig: velocity} provides an example of velocity learning from IMU sequence, where the periodicity of human walking makes it easy to infer the user's moving velocity. Similar observations have been made for vehicles, UAVs, and robotic platforms, which will be discussed in Section \ref{sec: vehicles}. Existing works on applying learned velocity to correct IMU integration can generally be divided into three categories, as shown in Figure \ref{fig: integration}, and will be discussed as follows.

One category of deep learning models aims to learn location displacement, which is the average velocity multiplied by a fixed period of time, as illustrated in Figure \ref{fig: integration}(a). The approach proposed by \cite{chen2018ionet} formulates inertial positioning as a sequential learning problem, where 2D motion displacements in the polar coordinate, also known as polar vectors, are learned from independent windows of segmented inertial data. This is because the frequency of platform vibrations is relevant to the absolute moving speed, which can be measured by IMU, when tracking human or wheeled configurations. Based on this observation, they propose IONet, an LSTM-based framework for end-to-end learning of relative poses. Trajectories are generated by adding motion displacements together with initial locations. To train neural models, a large collection of data was collected from a smartphone-based IMU in a room with a high-precision visual motion tracking system (i.e., Vicon) to provide ground-truth pose labels. Once the model is trained, the IONet model can be used in areas outside the data-collection room. In a two-minute random pedestrian walking scenario, the localization error of IONet is within 3 meters 90\% of the time, when evaluating across users, devices, and attachments, outperforming some classical PDR algorithms. In tracking trolley, IONet shows comparable performance over representative visual-inertial odometry and is even more robust in featureless areas. However, supervised learning-based IONet requires high-precision pose as training labels. When testing with data different from those in the training set, there will be performance degradation. To improve the generalization ability, \cite{chen2019motiontransformer}  proposes MotionTransformer, which allows the inertial positioning model to self-adapt into new domains via generative adversarial network (GAN) \cite{goodfellow2020generative} and domain adaptation \cite{tzeng2017adversarial}, without the need for labels in new domains. To encourage more reliable inertial positioning, \cite{chen2019deep} is able to produce pose uncertainties along with poses, offering the belief in the extent to which the learned pose can be trusted. To allow full 3D localization, TLIO \cite{liu2020tlio}  proposes to learn 3D location displacements and covariances from a sequence of gravity-aligned inertial data. To avoid the impacts from initial orientation, the inertial data are transformed into a local gravity-aligned frame. The learned displacements and covariances are then incorporated into an extended Kalman filter as observation states that estimate full-states of orientation, velocity, location, and IMU bias. In a 3-7 minute human motion scenario, the localization error of TLIO is within 3 meters 90\% of the time.

Another category of deep learning models aims to leverage learned velocity to correct accelerations, as illustrated in Figure \ref{fig: integration}(b). A prominent example is RIDI \cite{yan2018ridi}, which trains a deep neural network to predict velocity vectors from inertial data, which are then used to correct linear accelerations by subtracting gravity, aligning with the constraints of learned velocities. The corrected linear accelerations are then doubly integrated to estimate positions. To enhance the accuracy of inertial accelerations, RIDI leverages human walking speed as a prior, which compensates for the drifts in inertial positioning, effectively constraining them to a lower level. RoNIN  \cite{herath2020ronin} improves upon RIDI by transforming inertial measurements and learned velocity vectors into a heading-agnostic coordinate frame and introducing several novel velocity losses. To minimize the impact of orientation estimation, RoNIN employs device orientation to transform inertial data into a frame with its Z-axis aligned with gravity. However, a limitation of RoNIN is its reliance on orientation estimation.
NILoc \cite{herath2022neural} is an intriguing trial based on RoNIN, which tackles the neural inertial localization problem, aiming to infer global location from inertial motion history only. This work recognizes that human motion patterns are unique in different locations, which can be utilized as a "fingerprint" to determine the location, similar to WiFi or magnetic-field fingerprinting. NILoc first calculates a sequence of velocity from inertial data and then employs a Transformer-based DNN framework \cite{vaswani2017attention} to transform the velocity sequence into location. However, one fundamental limitation of NILoc is that in some areas, such as open spaces, symmetrical or repetitive places, there may not be a unique motion pattern.

An alternative approach involves incorporating learned velocity into the updating process of a Kalman filter (KF), as shown in Figure \ref{fig: integration} (c). \cite{cortes2018deep} uses a ConvNet to infer current speed from IMU sequences and incorporates this speed into the Kalman filter as a velocity observation to constrain the drifts of SINS-based inertial positioning. This approach is similar to the zero-velocity update (ZUPT) method, which detects and uses zero-velocity in KF as observations, but instead uses full speeds as observations in KF. Incorporating learned velocity allows the KF to handle more complex human motion. A similar trial is \cite{wang2020pedestrian}, that is based on a DNN that infers walking velocity in the body frame and combines it with an extended KF. In addition to the learned velocity, \cite{wang2020pedestrian} produces a noise parameter for KF to dynamically update parameters, rather than setting a fixed noise parameter.

 Inertial positioning heavily relies on accurately estimating the device's attitude. Several methods aim to improve orientation estimation to enhance the performance of deep learning based inertial odometry. RIDI, RoNIN, and TLIO still depend on device orientation to rotate inertial data into a suitable frame. To address this problem, IDOL \cite{sun2021idol} proposes a two-stage process that first learns orientation from data and then rotates inertial data into the appropriate frame, followed by learning the position. \cite{wang2022magnetic} estimates orientation using magnetic data and combines it with learned odometry to reduce positioning drifts while minimizing reliance on device orientation.

\begin{figure}
    \centering
\includegraphics[width=0.5\textwidth]{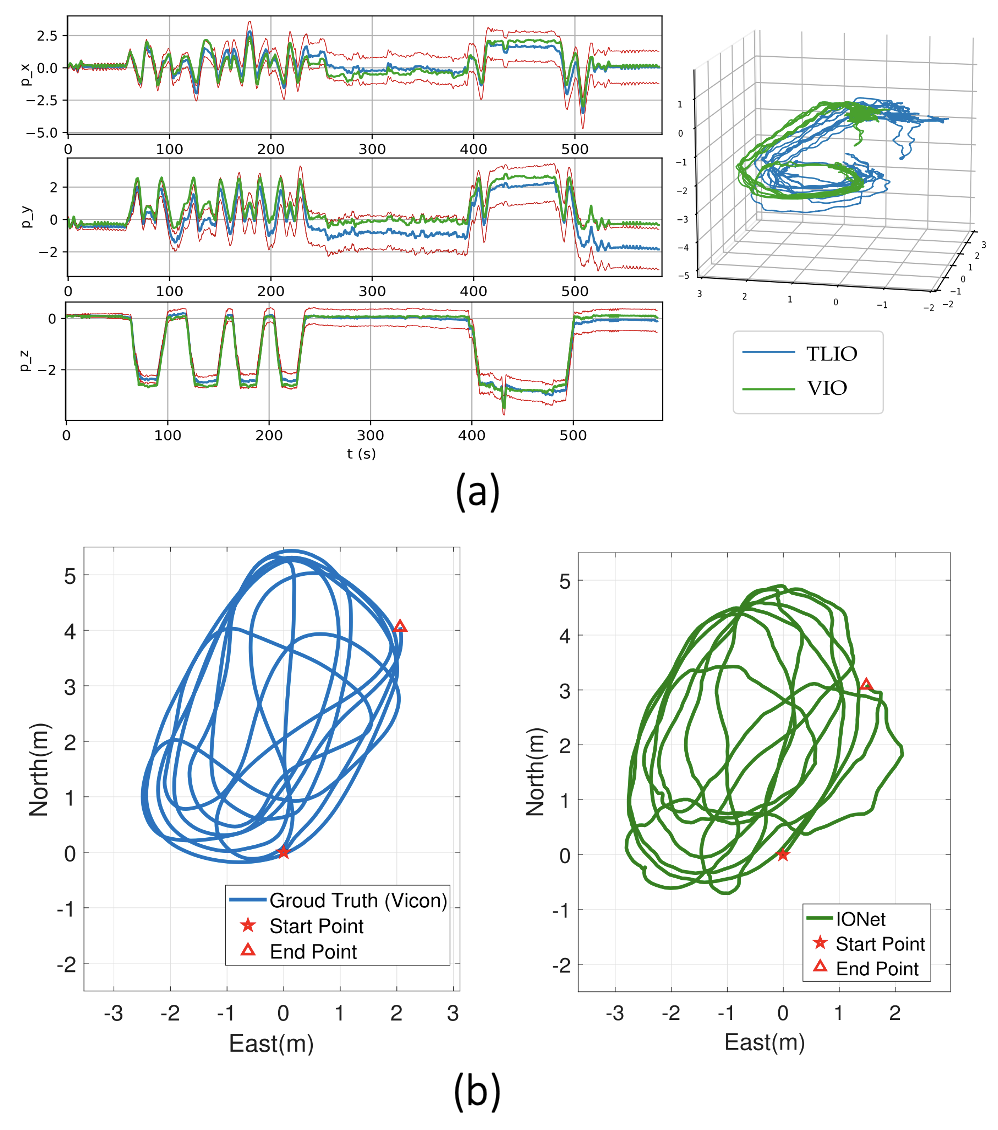}
    \caption{Sample results of deep learning based inertial positioning from (a) VR device for pedestrian tracking (reprint from TLIO \cite{liu2020tlio}) (b) smartphone for trolly tracking (reprint from IONet \cite{chen2018ionet}) }
    \label{fig: loc}
\end{figure}

Figure \ref{fig: loc} showcases several examples of deep learning based inertial positioning results.

\section{Learning Based IMU Integrated Positioning}
Integrating inertial sensors with other sensors as a multisensor navigation system has been an area of research for several decades. Nowadays, platforms such as robots, vehicles, and VR/AR devices are equipped with GNSS, cameras, IMUs, and LIDAR sensors. Hence, it is natural to consider introducing multimodal learning techniques \cite{ramachandram2017deep} and designing learning models capable of fusing multimodal information to construct a mapping function from sensor data to pose.

\subsection{Learning Based Visual-Inertial Positioning}

\begin{table*}[t]
    \caption{A summary of existing methods on deep learning based sensor fusion.}
    \label{tb: fusion}
    \small
    \begin{center}
    \begin{tabular}{c c c c c c}
 \hline
name & year & sensor & model & learning  & target \\
\hline
VINet\cite{Clark2017a} & 2017 & MC+I & ConvNet, LSTM  & SL & formulating VIO as a sequential learning problem \\
VIOLearner\cite{shamwell2019unsupervised} & 2018 & MC+I & ConvNet & UL &  VIO with online correction module \\
Chen et al.\cite{chen2019selective} & 2019 & MC+I & ConvNet, LSTM, Attention & SL & feature selection for deep VIO \\
DeepVIO\cite{han2019deepvio} & 2019 & SC+I & ConvNet, LSTM & UL & learning VIO from stereo images and IMU \\
DeepTIO\cite{saputra2020deeptio} & 2020 & T+I & ConvNet, LSTM, Attention & SL & learning pose from thermal and inertial data \\
MilliEgo \cite{lu2020milliego} & 2020 & MR+I & ConvNet, LSTM, Attention & SL & learning pose from mmWare radar and inertial data\\ 
UnVIO \cite{wei2021unsupervised} & 2021 & MC+I & ConvNet, LSTM, Attention & UL & unsupervised learning of VIO \\
DynaNet \cite{chen2021dynanet} & 2021 & MC+I & ConvNet, LSTM & SL & combining DNN with Kalman filtering  \\
SelfVIO \cite{almalioglu2022selfvio} & 2022 & MC+I & ConvNet, LSTM, Attention & UL & unsupervised VIO with GAN-based depth generator \\
Tu et al. \cite{tu2022undeeplio} & 2022 & L+I & ConvNet, LSTM, Attention & UL & unsupervised learning of LIDAR-inertial odometry\\
\hline
\end{tabular}
\end{center}
\begin{itemize}
    \footnotesize{
    \item \textit{Year} indicates the publication year of each work.
                \item \textit{Sensor} indicates the sensors involved in each work. I, MC, SC, T, MR, L, A represent inertial sensor, monocular camera, stereo camera, thermal camera, millimeter wave radar, LIDAR and airflow sensor respectively.
                \item \textit{Learning} indicates how to train neural networks. SL and UL represent Supervised Learning and Unsupervised Learning.
    }
    \end{itemize}
\end{table*}

\begin{figure}
    \centering
    \includegraphics[width=0.5\textwidth]{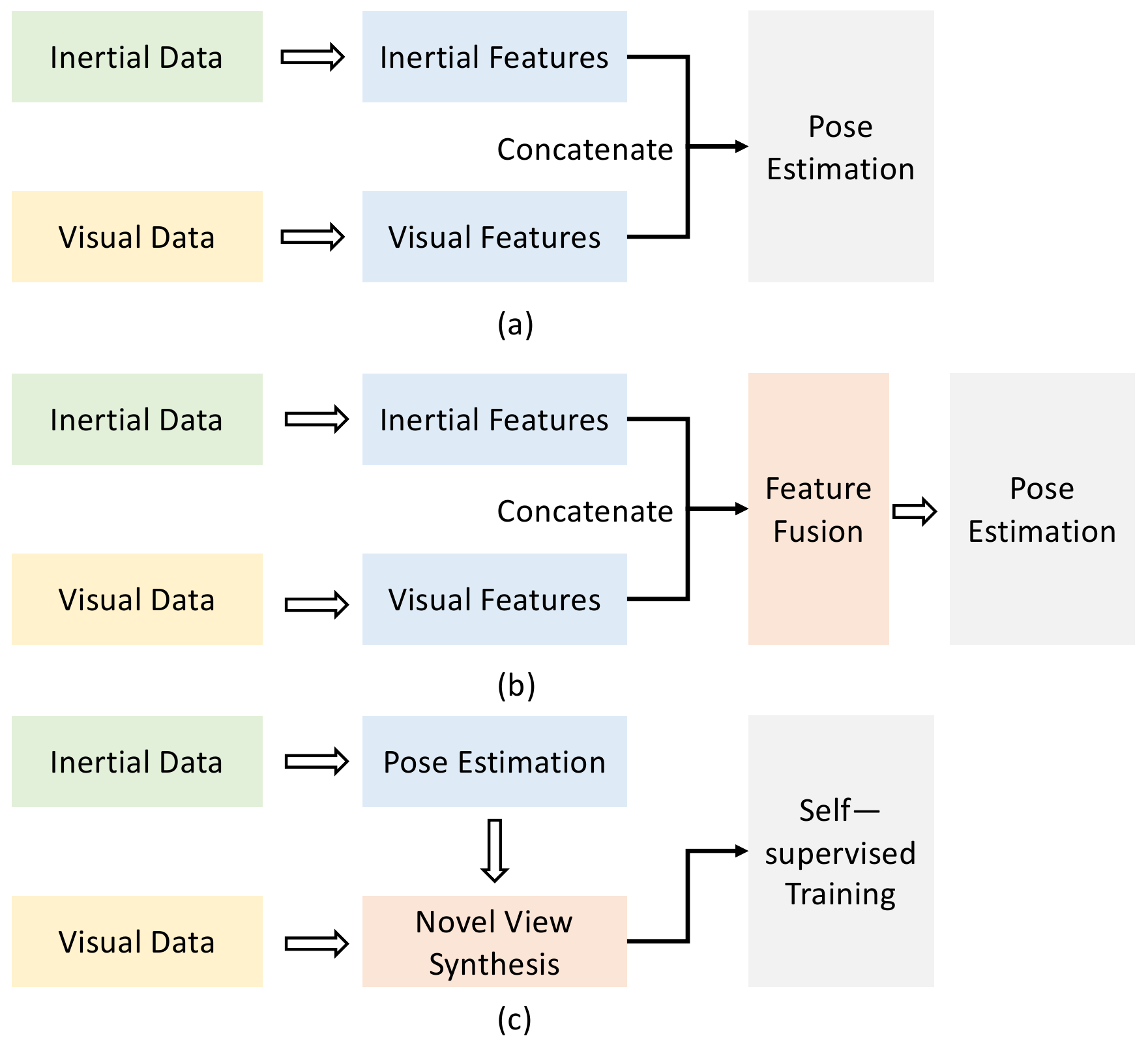}
    \caption{An overview of existing methods on deep learning based sensor fusion for visual-inertial positioning}
    \label{fig: fusion}
\end{figure}

Visual-inertial odometry (VIO) has garnered attention as a means of integrating low-cost, complementary camera and IMU sensors that are widely deployed. Monocular vision can capture the appearance and geometry of a scene, but cannot recover the scale metric. IMU provides metric scale and improves motion tracking in featureless areas, complex lighting conditions, and motion blur. However, a pure inertial solution can only last for a short period. Therefore, an effective fusion of these two complementary sensors is necessary for accurate pose estimation.

Traditional VIO methods integrate visual and inertial information based on filtering \cite{jones2011visual,Li2013b}, fixed-lag smoothing \cite{leutenegger2015keyframe}, or full smoothing \cite{forster2017manifold}. Recently, deep learning-based VIO models have emerged, directly constructing a mapping function from images and IMU to pose in a data-driven manner. VINet \cite{Clark2017a} is an end-to-end deep VIO model consisting of a ConvNet-based visual encoder to extract visual features from two images and an LSTM-based inertial encoder to extract inertial features from a sequence of inertial data between the two images. As shown in Figure \ref{fig: fusion} (a), the visual and inertial features are concatenated together as one tensor, followed by an LSTM and fully-connected layer that finally maps features into a 6-dimensional pose. VINet is trained on public driving datasets such as the KITTI dataset \cite{geiger2013vision} and a public drone dataset such as the EuroC dataset \cite{burri2016euroc}. The learned VIO model is generally more robust to sensor noises compared to traditional VIO methods, although its model performance still cannot compete with state-of-the-art VIO methods.

To effectively integrate visual and inertial information, \cite{chen2019selective} proposes a selective sensor fusion mechanism that learns to choose important features conditioned on sensor observations, as demonstrated in Figure \ref{fig: fusion} (b). Specifically, this work proposes two types of fusion: soft fusion, which is based on an attention mechanism and generates a soft mask to reweight features based on their importance, and hard fusion, which is based on Gumbel Soft-max and generates a hard mask consisting of either 1 or 0 to either propagate or ignore a feature. Experimental evaluation on the KITTI dataset demonstrates that compared with directly concatenating features \cite{Clark2017a}, selective fusion enhances the performance of deep VIO by 5\%-10\%. An interesting observation is that the number of useful features is relevant to the amount of linear/rotational velocity, with inertial features contributing more to rotation rate (e.g., turning), while more visual features are used to increase linear velocity.

Both \cite{Clark2017a} and \cite{chen2019selective} are trained in a supervised learning manner using datasets with high-precision ground-truth poses as training labels. However, obtaining high-precision poses can be difficult or costly in certain cases. Consequently, self-supervised learning-based VIOs, which do not require pose labels, have attracted attention. Self-supervised VIOs leverage the multi-view geometry relation of consecutive images, such as novel view synthesis, as a supervision signal \cite{shamwell2019unsupervised,han2019deepvio,wei2021unsupervised,almalioglu2022selfvio}. The task of novel view synthesis involves transforming a source image into a target view and comparing the differences between the synthesized target images and real target images as loss.
In VIOLearner \cite{shamwell2019unsupervised} and DeepVIO \cite{han2019deepvio}, as shown in Figure \ref{fig: fusion} (c), the pose transformation is generated from an inertial data sequence and used in the novel view synthesis process. In UnVIO \cite{wei2021unsupervised} and SelfVIO \cite{almalioglu2022selfvio}, inertial data is integrated with visual data via an attention module applied to the concatenated visual and inertial features extracted from the images and IMU sequence. They show that incorporating inertial data with visual data improves the accuracy of pose estimation, particularly rotation estimation.

\subsection{Learning based IMU/GNSS Integrated Positioning}

Kalman filtering (KF) serves as a conventional solution for merging IMU and GPS data in GNSS/IMU integrated navigation systems. However, recent developments leverage deep learning to enhance the accuracy of filtering algorithms and mitigate IMU/GNSS positioning drifts.

\cite{jwo2012neural} introduces a tight-integration of GNSS and INS by converting INS information into Doppler data and integrating it with GNSS tracking loops to mitigate Doppler effects on the GNSS signal. It incorporating radial basis function neural network and Adaptive Network-Based Fuzzy Inference System, effectively bridges GNSS outages, contributing to overall system robustness. After that, \cite{li2014gps} introduces a GNSS/INS/odometer integrated system for land vehicle navigation, utilizing a fuzzy neural network (FNN) to refine resolution accuracy during prolonged GNSS outages. Their experimental results validate its effectiveness in improving position, velocity, and attitude accuracy, particularly under extended GPS signal loss.

Moving beyond the neural network methods before deep learning age, \cite{hosseinyalamdary2018deep} proposes a deep learning-based Kalman filtering approach. It incorporates a modeling step alongside the prediction and update steps of the extended KF, addressing IMU errors and correcting positioning drifts with the help of GNSS measurements. Expanding on the dual optimization concept, \cite{shen2019dual} introduces two neural networks to optimize INS/GNSS navigation during GNSS outages. The first network compensates for inertial navigation system drifts, while the second corrects errors generated by a filtering process, employing a radial basis function network for accurate position data.

In scenarios where GNSS signals are absent, \cite{lu2020heterogeneous} proposes a multi-task learning method. It involves denoizing inertial data through a convolutional autoencoder, followed by temporal convolutional network (TCN) processing to address GNSS gaps. The resulting aiding data significantly contributes to deriving an accurate navigation solution within the Kalman filtering (KF) framework. \cite{shen2020seamless} introduces the Self-Learning Square-Root Cubature Kalman Filter (SL-SRCKF). This method employs an LSTM-based framework to continuously obtain observation vectors during GNSS outages, learning the relationship between observation vectors and internal filter parameters. The SL-SRCKF's error prediction ability is notably enhanced by introducing long short-term memory (LSTM) network, outperforming other neural networks under various GPS outage conditions. 

To refine the parameters used in filtering, \cite{wu2020predicting} introduces temporal convolutional neural (TCN) based framework that not only learns the parameters of measurement noise covariances but also the parameters of process noise covariances, resulting in higher position accuracy compared to classical INS/GNSS integrated positioning systems. Additionally, \cite{xiao2021residual} proposes a residual neural network with an attention mechanism to predict individual velocity elements of the noise covariance matrix. Through experiments, this work demonstrates that adjusting the non-holonomic constraint uncertainty during dynamic vehicle motions improves positioning accuracy, particularly under large dynamic motions.

\subsection{Learning to Fuse IMU with Other Sensors}
The use of learning-based sensor fusion extends beyond visual-inertial odometry (VIO) and IMU/GNSS integrated navigation to include other sensor modalities such as LiDAR-inertial odometry (LIO), thermal-inertial odometry, and radar-inertial odometry \cite{tu2022undeeplio,saputra2020deeptio,lu2020milliego}. DeepTIO \cite{saputra2020deeptio} and MilliEgo \cite{lu2020milliego} employ attention-based selective fusion mechanisms, similar to soft fusion \cite{chen2019selective}, to reweight and fuse features from inertial and visual data, resulting in improved pose accuracy. In addition, unsupervised learning-based LiDAR-inertial odometry \cite{tu2022undeeplio} generates motion transformation from IMU sequence and uses it for LIDAR novel view synthesis to facilitate self-supervised learning of egomotion, similar to VIOLearner \cite{shamwell2019unsupervised}. In all these cases, the inclusion of IMU data in deep neural networks enhances pose estimation accuracy and robustness.

\section{Learning to Correct Pedestrian Inertial Positioning}
The previous subsections addressed the general problems of applying deep learning to correct inertial positioning drifts and sensor fusion for IMU integrated positioning. This subsection focuses on the specific use of deep learning to address particular aspects of pedestrian navigation algorithms, namely Pedestrian Dead Reckoning (PDR) and Zero-Velocity Update (ZUPT).

\subsection{Learning to Correct Pedestrian Dead Reckoning}
Pedestrian dead reckoning (PDR) error drifts often stem from inaccurate stride and heading estimates. To address these issues, researchers have incorporated deep learning techniques into the process of step detection, dynamic step length estimation, and walking heading estimation.

To estimate walking stride more robustly, researchers have sought to solve it in a data-driven way. One such method is SmartStep \cite{Abiad2022}, a deep learning-based step detection framework that achieves 99\% accuracy in step detection tasks across various motion modes. Compared to peak/valley detection-based methods, data-driven methods do not require IMUs to be fixed in position, specific motion modes, or pre-calibration and threshold setting. Another approach involves using LSTM to regress walking stride from raw inertial data \cite{feigl2019bidirectional}. This method has demonstrated effectiveness in various human motions, such as walking, running, jogging, and random movements. Additionally, StepNet \cite{klein2020stepnet} learns to estimate step length dynamically, i.e., the change in distance, which achieves an impressive performance with only a 2.1\%-3.2\% error rate when compared to traditional static step length estimation. The attachment mode of the device, such as in hand or in pocket, can also influence walking stride estimation. To address this problem, Bo et al. \cite{bo2022mode} employed domain adaptation \cite{tzeng2017adversarial} to extract domain-invariant features for stride estimation, which enhanced the performance in new domains, such as holding, calling, pocket, and swinging.

 Accurate heading estimation is crucial for updating position in the right direction in PDR. To achieve more accurate and robust heading estimation, Wang et al. \cite{wang2019pedestrian} utilize a Spatial Transformer Network \cite{jaderberg2015spatial} and LSTM to learn heading direction from the inertial sensor attached to an unconstrained device. However, one problem that arises is the misalignment between the device heading and pedestrian heading, making it difficult to estimate the real walking heading based on sensor data.
To address this misalignment issue, \cite{Manos2022} introduces a deep neural network to estimate walking direction in the sensor's frame. They derive a geometric model to convert walking direction from the sensor's frame into a reference frame (i.e., north and east coordinates) by exploiting acceleration and magnetic data. This geometric model is combined with a learning framework to produce heading estimates. When tested on unseen data, this work reports a median heading error of 10\degree.
PDRNet \cite{asraf2021pdrnet} follows the process of a traditional PDR algorithm but replaces the step length and heading estimation modules with deep neural networks. Their experiments indicate that learning step length and heading together outperforms regressing them separately.

\subsection{Learning to Correct Zero-Velocity Update}
In pedestrian inertial navigation systems (INS) based on zero-velocity update (ZUPT), the zero-velocity phase is utilized to correct inertial positioning errors through Kalman filtering. Therefore, the accuracy of zero-velocity detection is crucial in determining when to update the system states. However, traditional threshold-based zero-velocity detection is complicated by the mixed variety of motions experienced by humans, making it challenging to set a reliable threshold when the user is still.

To address this issue, researchers have explored data-driven approaches that utilize the powerful feature extraction and classification capabilities of deep learning to classify whether the user is in the ZUPT phase. For instance, \cite{wagstaff2018lstm} proposes a six-layer long short-term memory (LSTM) network to detect zero-velocity. The LSTM inputs a sequence of IMU data, typically 100 consecutive data points, and outputs the probability of whether the user is still or in motion at the current timestep. The results from the LSTM-based zero-velocity detection are then fed into a ZUPT-based INS.
The proposed approach achieves a reduction in localization error by over 34\% compared to fixed threshold-based ZVDs and was shown to be more robust during a mixed variety of motions, such as walking, running, and climbing stairs. Similarly, \cite{yu2019azupt} designs an adaptive ZUPT using convolutional neural networks (ConvNet) to classify ZVDs based on IMU sequences. Deep learning approaches, such as LSTM and ConvNet, have demonstrated excellent performance in extracting robust and useful features for zero-velocity identification, irrespective of different users, motion modes, and attachment places.

\section{Learning to Correct Inertial Positioning on Vehicles, UAV and robotic platforms}
\label{sec: vehicles}
As previously mentioned, deep learning methods have shown great potential in addressing the challenges of pedestrian inertial navigation. However, these techniques can also be applied to other platforms, such as vehicles, UAVs, robots, and more.

These platforms share similarities with pedestrians, such as the ability to infer movement velocity from inertial data. This is because inertial data contains vibration information that reflects the fundamental frequency proportional to the vehicle speed. Building on the success of IONet \cite{chen2018ionet}, \cite{esfahani2019aboldeepio} proposes AbolDeepIO, an improved triple-channel LSTM network that predicts polar vectors for drone localization from inertial data sequences. AbolDeepIO has been evaluated on a public drone dataset and has shown competitive performance compared to traditional visual-inertial odometry methods like VINS-mono. 

When deploying deep learning-based inertial navigation on real-world devices, prediction accuracy and model efficiency must be considered.
To address this, TinyOdom \cite{saha2022tinyodom} aims to deploy neural inertial odometry models on resource-constrained devices. It proposes a lightweight model based on temporal convolutional networks (TCN) \cite{lea2017temporal} to learn position displacement and optimizes the model through neural architecture search (NAS) \cite{ren2021comprehensive} to reduce model size between 31 and 134 times. TinyOdom was extensively evaluated on tracking pedestrians, animals, aerial, and underwater vehicles. Within 60 seconds, its localization error is between 2.5 and 12 meters.

Learning-based inertial odometry has also been extended to legged robots by \cite{buchanan2022learning}. The learned location displacement is combined with kinematic motion models to estimate robot system states at high frequencies (400 Hz). In this work, the robot successfully navigated a field experiment, where a legged robot walked around for 20 minutes in a mine with poor illumination and visual feature tracking failures.

In the realm of inertial positioning for vehicles, researchers have proposed various methods to mitigate error drifts and improve accuracy. One such method is presented in \cite{brossard2020ai}, where error covariances are learned from inertial data and incorporated into Kalman filtering for updating system states. This approach has been shown to improve inertial positioning performance.
Similar to ZUPT-based pedestrian positioning, zero-velocity-update (ZUPT) can also be used for car-equipped inertial navigation systems. The zero-velocity phase provides valuable context information to correct system error drifts via Kalman filtering. OdoNet, presented in \cite{tang2022odonet}, is an example of a system that learns and utilizes car speed along with a zero-velocity detector to reduce error drifts in car-equipped IMU systems.
Deep learning techniques have also been explored for detecting zero-velocity phases in vehicle navigation. For example, \cite{brossard2019rins} proposes a deep learning-based method for detecting zero-velocity phases in vehicle navigation. In another study, \cite{zhang2021imu} derives a model with motion terms that are relevant only to the IMU data sequence. This model provides theoretical guidance for learning models to infer useful terms and has been evaluated on a drone dataset, where it outperformed TLIO and other learning methods.

Overall, these studies demonstrate the potential of deep learning-based methods in improving inertial navigation for various platforms, including pedestrians, vehicles, drones, and robots. By leveraging the rich information contained within IMU data, deep learning models can effectively mitigate error drifts and improve the accuracy of inertial positioning systems. Furthermore, by optimizing the model efficiency and considering deployment on resource-constrained devices, these techniques can be applied in real-world scenarios.


\begin{table*}[ht]
\caption{Public Datasets for Data-Driven Inertial Positioning}
\label{tb: dataset_compare}
\begin{center}
\begin{tabular}{c c c c c c c c c}
\hline
Dataset & Year & Environment & Attachment & IMU Type & Sample Rate & Groundtruth &  Accuracy & Data Size\\
\hline
KITTI Odometry & 2013 & Outdoors & Car & OXTS RT3003 & 10 Hz & GPS/IMU & 10 cm & 22 seqs, 39.2 km\\
EuRoC MAV & 2016 & Indoors & MAV & ADIS 16488 & 200 Hz & Motion Capture & 1 mm & 11 seqs, 0.9 km\\
Oxford RobotCar & 2016 & Outdoors & Car & NovAte SPAN & 50 Hz & GPS/IMU & Unknown & 1010.46 km\\
TUM VI & 2018 & In/Outdoors & Human & BMI 160 & 200 Hz & Motion Capture & 1 mm & 28 seqs, 20 km\\
ADVIO & 2018 & In/Outdoors & Human & InvenSense 20600 & 100 Hz & Other Algorithms & Unknown & 23 seqs, 4.5 km\\
OxIOD & 2018 & Indoors & Human & InvenSense 20600 
& 100 Hz & Motion Capture & 0.5 mm & 158 seqs, 42.587 km\\
RONIN & 2019 & Indoors & Human & - & 200 Hz & AR device & Unknown & 117 seqs\\
SIMD & 2023 & In/Outdoors & Human & - & 50 Hz & GPS/IMU & 10 cm & 4562 seqs, 717.48 km\\
\hline
\end{tabular}
\end{center}
\end{table*}

\section{Learning Sensor Fusion for IMU integrated Positioning Applications}
This section explores the applications of learning-based sensor fusion for IMU-integrated positioning in vehicles, robots, and pedestrian navigation. Compared to pure inertial positioning, IMU-integrated systems demonstrate enhanced robustness and accuracy in complex dynamic scenarios, enabling sustained operation over extended periods.

In the realm of vehicle navigation, establishing a robust tight-integrated IMU/GNSS positioning system is crucial for providing accurate global positioning, particularly in challenging environments like tunnels or streets with tall buildings. Deep learning techniques play a pivotal role in addressing challenges such as compensating positioning drifts and estimating filtering parameters. Studies like \cite{hosseinyalamdary2018deep, shen2019dual} utilize deep learning to model error drifts in IMU/GNSS systems, while others such as \cite{wu2020predicting, xiao2021residual} focus on learning essential filtering parameters like measurement noises, process noises, or velocity for effective fusion in Kalman filtering.

In the domain of intelligent unmanned platforms like vehicles, robots, and UAVs, which often operate in complex and dynamic scenarios, robust perception is vital for reliable planning, decision-making, and control. Multisensory positioning supports these objectives, but issues such as camera occlusions, image degradations, and complex illumination changes can make visual-inertial positioning systems fragile. Deep learning interventions, as seen in studies like \cite{Clark2017a, chen2019selective}, enhance the robustness of visual-inertial positioning by extracting more reliable features through the efficacy of deep neural networks in feature learning. End-to-end training in self-supervised learning-based visual-inertial positioning \cite{shamwell2019unsupervised, han2019deepvio, wei2021unsupervised} leverages multi-view geometry relations between consecutive images, providing a supervision signal without requiring high-precision pose labels. This self-supervised approach maximizes the use of large amounts of data, simultaneously offering depth estimates crucial for scene perception in self-driving vehicles and mobile robots.

For pedestrian navigation in indoor environments, vision or LiDAR-aided inertial positioning corrects drifts by exploiting feature associations between images or point clouds. However, this integrated system demands more energy and computation compared to IMU-only solutions. To enhance the efficiency of deep learning-based visual-inertial positioning for pedestrians, \cite{saputra2019distilling} employs knowledge distillation to compress a large teacher network into a lightweight version, transferring learned knowledge effectively. In scenarios with significant image degradations, such as smoke or fog, integrating thermal or mmWave radar sensors with IMU proves beneficial. Approaches like DeepTIO \cite{saputra2020deeptio} and MilliEgo \cite{lu2020milliego} utilize generative model-based frameworks to extract features from thermal images or noisy mmWave radar point clouds, constructing multisensory positioning systems for accurate pose estimation, particularly in smoke-filled environments for firefighters.

\section{Learning based Human Motion analysis and Activity Recognition}
Inertial sensors have diverse applications beyond positioning, such as motion tracking, activity recognition, and more. Although these tasks are not the primary focus of this survey, this section provides a brief yet comprehensive overview of how deep learning is utilized in these domains.

\subsection{Human Motion Analysis}
Data-driven approaches are utilized to reconstruct human pose and motion using either a single IMU or multiple IMUs attached to the body. These models primarily focus on analyzing human motion rather than localizing users, which differentiates them from inertial positioning. Several studies have applied machine learning to gait and pose analysis, such as knee angle estimation for human walking using supervised support vector regression in \cite{Ahuja2011} and probabilistic parameter learning for human gesture recognition in \cite{Parate2014} through handcrafted motion features extracted from inertial data. In addition, machine learning methods, such as multi-layer perceptrons (MLPs), have been utilized in IMU data to learn sensor displacement for human motion reconstruction in \cite{mannini2010machine,valtazanos2013using,yuwono2014unsupervised}.

Recently, deep learning has shown promising performance in human pose reconstruction. For example, \cite{huang2018deep} proposed Deep Inertial Poser, a recurrent neural network (RNN)-based framework that can reconstruct full-body pose from six IMUs attached to the user's body. TransPose \cite{yi2021transpose}, another RNN-based framework, enables real-time human pose estimation using six body-attached IMUs. Furthermore, \cite{yi2022physical} combines a neural kinematics estimator with a physics-aware motion optimizer to improve the accuracy of human motion tracking.

\subsection{Human Activity recognition (HAR)}
Deep learning can be utilized to exploit inertial information from body-worn IMUs for human activity recognition. For instance, \cite{anguita2013public} published a popular public dataset of human activity recognition and successfully classified current activity among six classes, including walking, standing still, sitting, walking downstairs, walking upstairs, and laying down, using support vector machines (SVM). In addition, \cite{chevalier2016lstms} presents an LSTM-based HAR model that inputted a sequence of inertial data and outputted class probability. Moreover, \cite{zebin2016human} introduces a ConvNet-based HAR model that achieved a classification accuracy of 97\%, outperforming an accuracy of 96\% from SVM-based HAR models. To reduce onboard computational requirements, \cite{ravi2016deep} presents a learning framework that exploited both features automatically extracted by DNN and hand-crafted features to achieve accurate and real-time human activity recognition on low-end devices.

Learning from inertial data can also benefit sports and health applications. For instance, \cite{eskofier2016recent} shows that deep learning is effective in detecting Parkinson's disease by assessing the patient's daily activity through the analysis of inertial information from wearable sensors. Additionally, \cite{windau2019inertial} provides instructions for athletes' sports training based on sensor data and activity information.

\section{Datasets and Evaluation Metrics}
In this section, we present five prominent public datasets widely utilized in deep learning-based inertial positioning research, as outlined in Table \ref{tb: dataset_compare}. Additionally, we introduce the evaluation metrics and compare several representative methods using two well-known datasets.

\subsection{The Inertial Positioning Datasets}
In the realm of vehicle navigation, the KITTI dataset \cite{geiger2013vision} serves as a widely adopted benchmark. The sensors are rigidly affixed to the car chassis, making it conducive for studying vehicle movements. Specifically, the KITTI Odometry Dataset encompasses data collected from car-driving scenarios, including visual images, LIDAR point clouds, IMU, and ground truth. The high-precision GPS/IMU integrated system provides ground truth, with raw unsynchronized data packages containing high-frequency inertial data at 100 Hz, and images and ground truth from GPS at 10 Hz.

In the domain of drone and robotic research, the EuRoC MAV datasets \cite{burri2016euroc} feature tightly synchronized video streams from a Micro Aerial Vehicle (MAV) equipped with a stereo camera and an IMU. Comprising 11 flight trajectories across two environments, this dataset captures complex motion. The images, captured at 20 frames per second (FPS), and IMU measurements recorded at 200 Hz span the MAV's stationary state, takeoff, flight, and landing on its initial position. A consistent IMU (MEMS IMU ADIS16448) operating at 200 Hz, along with Vicon Motion Capture and Leica MS50 laser tracker, is used to produce accurate ground truth.

For pedestrian navigation, the Oxford Inertial Odometry Dataset (OxIOD), The Robust Neural Inertial Navigation Dataset (RONIN), and Smartphone Inertial Measurement Dataset (SIMD) collect IMU data from mobile devices to reflect human motion in everyday life.
\begin{itemize}
    \item The OxIOD dataset \cite{chen2020deep} consists of inertial measurements collected with IMUs attached in various ways (handheld, in the pocket, in a handbag, and on a trolley/stroller). It encompasses different motion modes, four types of off-the-shelf consumer phones, and data from five users. With 158 sequences, the dataset covers a total walking distance and recording time of 42.5 km and 14.72 h.
    \item The RONIN dataset \cite{herath2020ronin}, containing inertial motion data from 100 human subjects, enables users to handle smartphones naturally as in day-to-day activities. It supports unrestricted phone orientation and placement, presenting a challenging task for developing inertial models invariant to device orientation or placement. Additionally, trajectories in the RONIN dataset have a larger spatial span compared to those in the OxIOD dataset.
    \item The SIMD dataset \cite{liu2023smartphone} encompasses over 4500 walking trajectories, totaling approximately 190 hours and covering more than 700 km. It includes diverse scenarios in four cities, both indoors and outdoors, seven phone attitudes, and involves more than 150 volunteers with their smartphones. The inertial data, collected by embedded smartphone IMU sensors, have synchronized timestamps and include specific force, angular rates, magnetic fields, and GPS-derived location information. Movement readings are also calculated by internal algorithms.
\end{itemize}


\subsection{Evaluation Metrics and Results}
To assess the efficacy of inertial positioning, researchers commonly employ two evaluation metrics: Absolute Trajectory Error (ATE) and Relative Trajectory Error (RTE) in the domain of deep learning-based inertial positioning.

\begin{itemize}
    \item Absolute Trajectory Error (ATE): ATE is quantified as the average root-mean-squared-error (RMSE) between the actual and predicted locations throughout the entire trajectory. A lower ATE signifies superior performance.
    \item Relative Trajectory Error (RTE): RTE is determined as the average root-mean-squared-error (RMSE) between the actual and predicted locations within a specified time interval. A lower RTE indicates more accurate predictions.
\end{itemize}

Here, we leverage the OxIOD and RoNIN datasets, which are widely used for pedestrian inertial navigation. Following the methods and results outlined in \cite{saha2022tinyodom}, we selected two traditional inertial positioning solutions, namely PDR and SINS, along with three representative learning-based models—IONet \cite{chen2018ionet}, RoNIN \cite{herath2020ronin}, and TinyOdom \cite{saha2022tinyodom} — to compare their performance on these datasets. The performance comparison is summarized in Table \ref{tb: performance}, where ATE and RTE values for each model on both datasets are presented.

PDR, a mainstream classical solution for pedestrian inertial navigation, employs a threshold-based step detector based on accelerometer peaks. Displacement is updated via Weinberg's stride length estimation model, and the heading is computed from the gyroscope readings. Traditional strapdown inertial navigation systems (SINS) integrate IMU measurements directly into position, velocity, and orientation based on Newtonian mechanics. However, due to error propagation from measurement noises, SINS quickly drifts and fails to provide reasonable results.
IONet regresses heading and location displacement from data, demonstrating superior performance in relative trajectory estimation on both datasets. Notably, PDR exhibits good performance on the OxIOD dataset, characterized by simpler and smoother pedestrian trajectories. However, IONet outperforms PDR on the RoNIN dataset, highlighting the effectiveness of learning-based motion modeling in complex scenarios.
RoNIN employs a heading-agnostic coordinate frame aligned with gravity, assuming correct orientation for learning inertial motion. In comparison, IONet and PDR do not make such an assumption. Consequently, RoNIN outperforms IONet and PDR significantly.
TinyOdom represents a lightweight learning-based inertial positioning model with significantly fewer neural network weights than RoNIN. While TinyOdom's performance is comparable to RoNIN on the OxIOD dataset, it lags behind RoNIN on the RoNIN dataset.

\begin{table}[t]
\begin{center}
\caption{The performance comparison of representative traditional and learning based inertial positioning solutions on two public datasets. Metric data is from \cite{saha2022tinyodom}.}
\label{tb: performance}
\begin{tabular}{ccccc}
\hline
\multicolumn{1}{l}{} & \multicolumn{2}{c}{OxIOD dataset}           & \multicolumn{2}{c}{RONIN dataset}           \\
\hline
\multicolumn{1}{l}{} & ATE (m)              & RTE (m)              & ATE (m)              & RTE (m)              \\
\hline
PDR                  & 3.47                 & 3.24                 & 34.81                & 23.62                \\
SINS                  & 9119.50              & 247.53               & 12398.00             & 59.85                \\
IONet                & 5.95                 & 2.84                 & 22.52                & 7.63                 \\
RoNIN                & 1.95                 & 0.42                 & 4.73                 & 1.21                 \\
TinyOdom             & 2.80                 & 1.26                 & 27.36                & 5.84                 \\
\hline
\end{tabular}
\end{center}
\end{table}

\section{Conclusions and Discussions}
In recent years, there has been a growing interest in using deep learning to address the problem of inertial positioning. This article provides a comprehensive review of the area of deep learning-based inertial positioning. The rapid advances in this field have already provided promising solutions to address problems such as inertial sensor calibration, the compensation of error drifts in inertial positioning, and multimodal sensor fusion. This section concludes and discusses the benefits that deep learning can bring to inertial navigation research, analyzes the challenges that existing research faces, and highlights the future opportunities of this evolving field.

\subsection{Benefits}
Unlike traditional geometric or physical inertial positioning models, the integration of deep learning into inertial positioning has led to the development of a range of alternative solutions to address the issue of positioning error drifts. The corresponding benefits can be summarized as follows:

\subsubsection{Learn to approximate complex and varying function}
The deep neural network has proven to be a powerful and versatile nonlinear function that can approximate the complex and variable factors involved in inertial positioning, which are difficult to model manually. For example, when calibrating sensors, the corrupt noises that exist in inertial measurements can be modeled and eliminated in a data-driven way by training on a large dataset using a DNN. Deep learning can also directly generate absolute velocity and position displacement from data, without the need for IMU integration, thus reducing positioning drifts. In pedestrian dead reckoning (PDR), deep learning can estimate step length based on data, rather than empirical equations, and implicitly remove the effects of different users. These works demonstrate that using a large dataset to build a data-driven model can produce more accurate motion estimates, as well as reduce and constrain the rapid error drifts of inertial navigation systems.

\subsubsection{Learn to estimate parameters}
Automatic identification of parameters through data-driven models contributes to paving the way for next-generation intelligent navigation systems that can actively exploit input data and evolve over time without human intervention. In classical inertial navigation mechanisms, certain parameters or modules need to be manually set and tuned before use. For instance, experts with experience need to settle parameters in Kalman filtering, such as observation noise, covariance, and process noise. Deep learning has proven effective in automatically producing suitable parameters for Kalman filtering based on input data \cite{brossard2020ai,chen2021dynanet,wang2019pedestrian}. In sensor calibration, reinforcement learning algorithms are used to discover optimal parameters for inertial calibration algorithms \cite{nobre2019learning}. In ZUPT-based pedestrian inertial positioning, deep learning is a viable solution for classifying zero-velocity phases and determining when to update system states.

\subsubsection{Learn to self-adapt in new domains}
Unforeseen or ever-changing issues in new application domains, such as changes in motion mode, carrier, and sensor noise, can significantly impact the performance of inertial systems. Learning models offer opportunities for inertial systems to adapt to new changes and overcome these influential factors implicitly by discovering and exploiting the differences in data distributions between domains.
For instance, \cite{chen2019motiontransformer} leverages transfer learning to allow INS to extract domain-invariant features from data, maintaining localization accuracy when sensor attachment is changed. The introduction of self-supervised learning enables navigation systems to learn from data without high-precision pose as training labels, allowing unlabelled inertial data to be effectively used for model performance improvement. In visual-inertial odometry, \cite{shamwell2019unsupervised,han2019deepvio,wei2021unsupervised} introduce novel view synthesis as a supervision signal to train deep VIO in a self-supervised learning way.
This self-adaptation ability is promising for mobile agents to continuously improve their localization performance in new application scenes.

\subsection{Challenges and Opportunities}
Despite the impressive and promising results that deep learning has already offered in inertial positioning, there are still challenges in existing methods when they are applied and deployed in real-world scenarios. To overcome these limitations, several opportunities and potential research directions are discussed below.

\subsubsection{Generalization and Self-learning}
The generalization problem is a major concern for deep learning-based methods because these models are trained on one domain (i.e., training set) but need to be tested on other domains (i.e., testing set). The possible differences in data between domains can lead to a degradation of prediction performance. Although deep learning-based inertial navigation models have reported impressive results on the author's own datasets, these works have not been evaluated in comprehensive experiments during long-term operation and across various devices, users, and application scenes. Thus, it is challenging to determine the real performance of these models in open environments.
To address the generalization problem, new learning techniques such as transfer learning \cite{weiss2016survey}, lifelong learning, and contrastive learning \cite{tian2020makes} can be introduced into inertial positioning systems, which is a promising direction. For instance, in the future, by exploiting information from physical/geometric rules or other sensors (e.g., GNSS, camera), the learning-based inertial positioning model can be self-supervisedly trained and enable mobile agents to learn from data in a lifelong manner.

\subsubsection{Black-box and Explainability}
Deep neural networks have been criticized as being a 'black-box' model due to their lack of explainability and interpretability. As these models are often used to support real-world tasks, it is crucial to investigate what is learned inside deep nets before deploying them to ensure their safety and reliability. Despite the good results shown by deep learning models in estimating important terms such as location displacement, sensor measurement errors, and filtering parameters, these terms lack concrete mathematical models, unlike traditional inertial navigation.
To determine whether these terms are trustworthy, uncertainties should be estimated in conjunction with the inertial positioning method \cite{chen2019deep} and used as indicators for users or systems to understand the extent to which model predictions can be trusted. In future research, it is important to reveal the governing mathematical or physical models behind the learned inertial positioning neural model and identify which parts of inertial positioning can be learned by deep nets. Introducing Bayesian deep learning into inertial positioning is also a promising direction that could offer interpretability for model predictions \cite{kendall2017uncertainties}.

\subsubsection{Efficiency and Real-World Deployment}
When deploying deep positioning models on user devices, it is crucial to consider the consumption of computation, storage, and energy in system design, in addition to prediction accuracy. Compared to classical inertial navigation algorithms, DNN-based inertial positioning models have a relatively large computational and memory burden, as they contain millions of neural parameters that require GPUs for parallel training and testing. Therefore, online inference of learning models, especially on low-end devices such as IoT consoles, VR/AR devices, and miniature drones, requires lightweight, efficient, and effective models. To achieve this goal, neural model compression techniques, such as knowledge distillation \cite{gou2021knowledge}, should be introduced to discover the optimal neural structure that balances prediction accuracy and model size. \cite{chen2020deep} and \cite{tang2022odonet} have conducted initial trials on minimizing the model size of inertial odometry.
Moreover, safety and reliability are also crucial factors to consider. In the future, it is worth exploring the optimal structure of learning-based inertial positioning models, considering model performance, parameter size, latency, safety, and reliability for real-world deployment.

\subsubsection{Data Collection and Benchmark}
Deep learning models' performance relies heavily on data quality, including dataset size, diversity, and consistency between training and testing sets. Ideally, deep learning-based inertial positioning models should be trained on diverse data across users, platforms, motion dynamics, and sensors to improve generalization. However, acquiring such data can be costly and time-consuming, and obtaining accurate ground-truth labels can be challenging. Previous research has varied in training data, model parameters, and evaluation metrics, hindering fair comparisons. 
In visual navigation tasks, such as visual odometry/SLAM, the KITTI dataset \cite{geiger2013vision} is commonly used as a benchmark to train and evaluate learning-based VO models. However, although published datasets for inertial navigation exist \cite{chen2020deep,herath2020ronin}, there is still a lack of a common benchmark that is adopted and recognized by mainstream methods in inertial positioning. In the future, a widely adopted dataset and benchmark, covering a variety of application scenarios, will greatly benefit and foster research in data-driven inertial positioning.

\subsubsection{Failure Cases and Physical Constraints}
Deep learning has demonstrated its capability in reducing the drifts of inertial positioning and contributing to various aspects of inertial navigation systems, as discussed in Section \ref{sec: ins}. However, DNN models are not always reliable and may occasionally produce large and abrupt prediction errors. Unlike traditional inertial navigation algorithms that are based on concrete physical and mathematical rules, DNN predictions lack constraints, and the failure cases must be considered in real-world applications with safety concerns.
To enhance the robustness of DNN predictions, possible solutions include imposing physical constraints on DNN models or combining deep learning with physical models as hybrid inertial positioning models. By doing so, the benefits from both learning and physics-based positioning models can be leveraged.

\subsubsection{New Deep Learning Methods}
Machine/deep learning is one of the fastest growing areas of AI, and its advances have influenced numerous fields such as computer vision, robotics, natural language processing, and signal processing. There are significant opportunities for applying deep learning techniques to inertial navigation and analyzing their effectiveness and theoretical underpinnings.
In the future, new model structures such as transformer \cite{vaswani2017attention}, diffusion models \cite{saharia2022photorealistic}, and generative models \cite{goodfellow2020generative}, and new learning methods such as transfer learning, reinforcement learning, contrastive learning \cite{tian2020makes}, unsupervised learning, and meta-learning \cite{finn2017model}, all hold promise for enhancing inertial positioning systems. Furthermore, advances in other domains such as neural rendering \cite{mildenhall2021nerf} and voice synthesis \cite{oord2018parallel} may provide valuable insights into developing more effective inertial positioning systems. Therefore, incorporating these rapidly-evolving deep learning methods into inertial navigation will be a significant area of research in the future.

\subsubsection{Deep Sensor Fusion}
Sensor fusion faces challenges like diverse sensor data formats, temporal sync issues, and sensor calibration complexity. Real-time processing needs, adapting to dynamic environments, and limited labeled data for multi-sensory models add more complexity. Deep learning in sensor fusion improves accuracy by merging inertial sensor data with others, learning fusion strategies, synchronization, and calibration from data. It adapts and customizes continuously for better performance in various applications like vehicles, robots, pedestrians, and drones.

\subsubsection{Robustness and Reliability}
In practical applications, the challenges associated with handling unforeseen situations and ensuring reliability become particularly critical, especially in safety-critical domains such as autonomous vehicles. The learning models employed may encounter difficulties in adapting to extreme conditions, thereby introducing risks to robust and reliable positioning. To solve these problems, prospective solutions such as diversifying training data, implementing adversarial testing and new learning models can contribute to the overall reliability of positioning system. In addition, continuous monitoring, adaptive algorithms, and strict certification standards play pivotal roles in enhancing the overall trustworthiness of the positioning system.

{\small
\bibliographystyle{ieeetr}
\bibliography{reference}
}

\begin{IEEEbiography}
[{\includegraphics[width=1in,height=1.25in,clip,keepaspectratio]{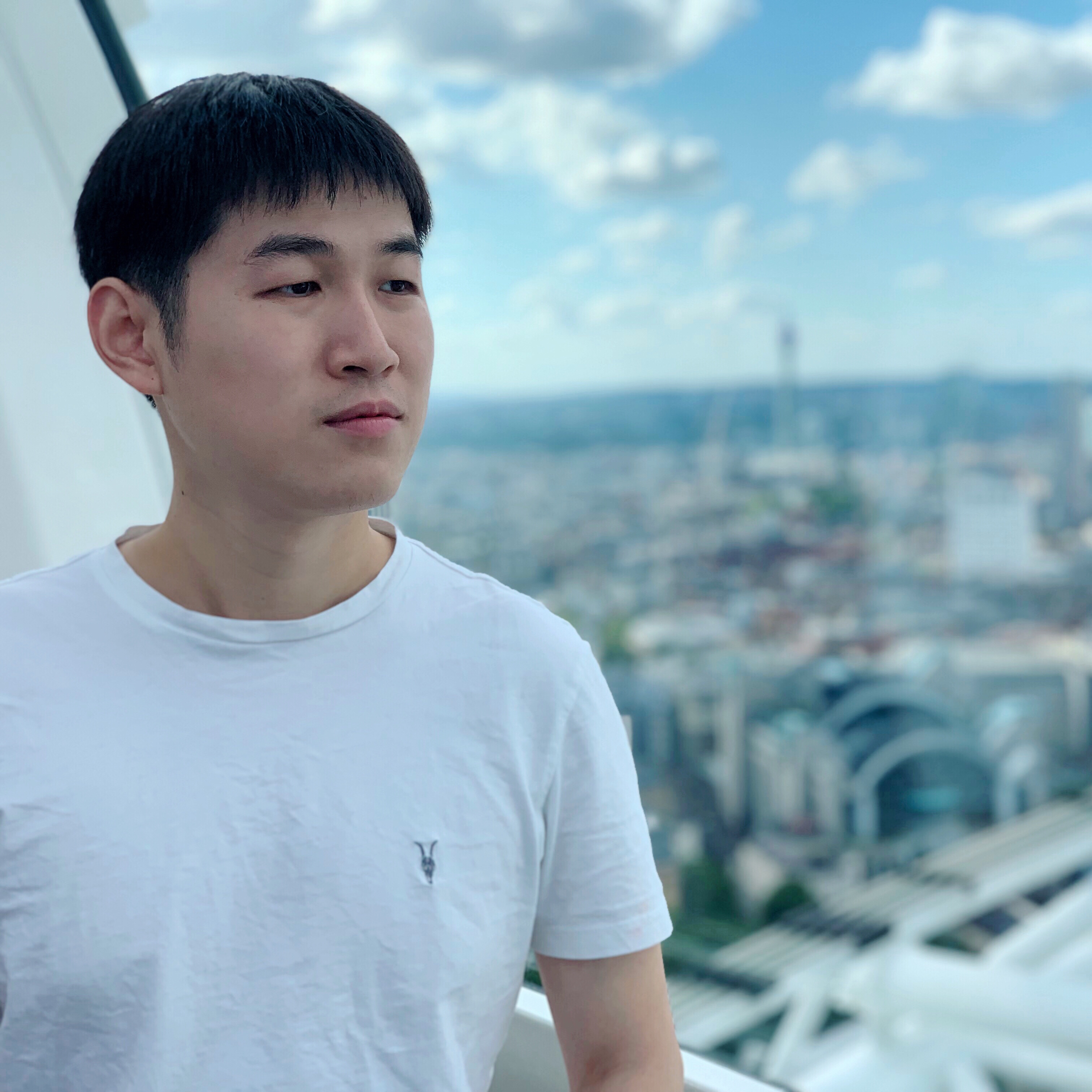}}]{Changhao Chen} obtained his Ph.D. degree at University of Oxford (UK), M.Eng. degree at National University of Defense Technology (China), and B.Eng. degree at Tongji University (China). Now he is an Assistant Professor at the College of Intelligence Science and Technology, National University of Defense Technology (China). His research interest lies in robotics, computer vision and cyber-physical systems. 
\end{IEEEbiography}

\begin{IEEEbiography}[{\includegraphics[width=1in,height=1.25in,clip,keepaspectratio]{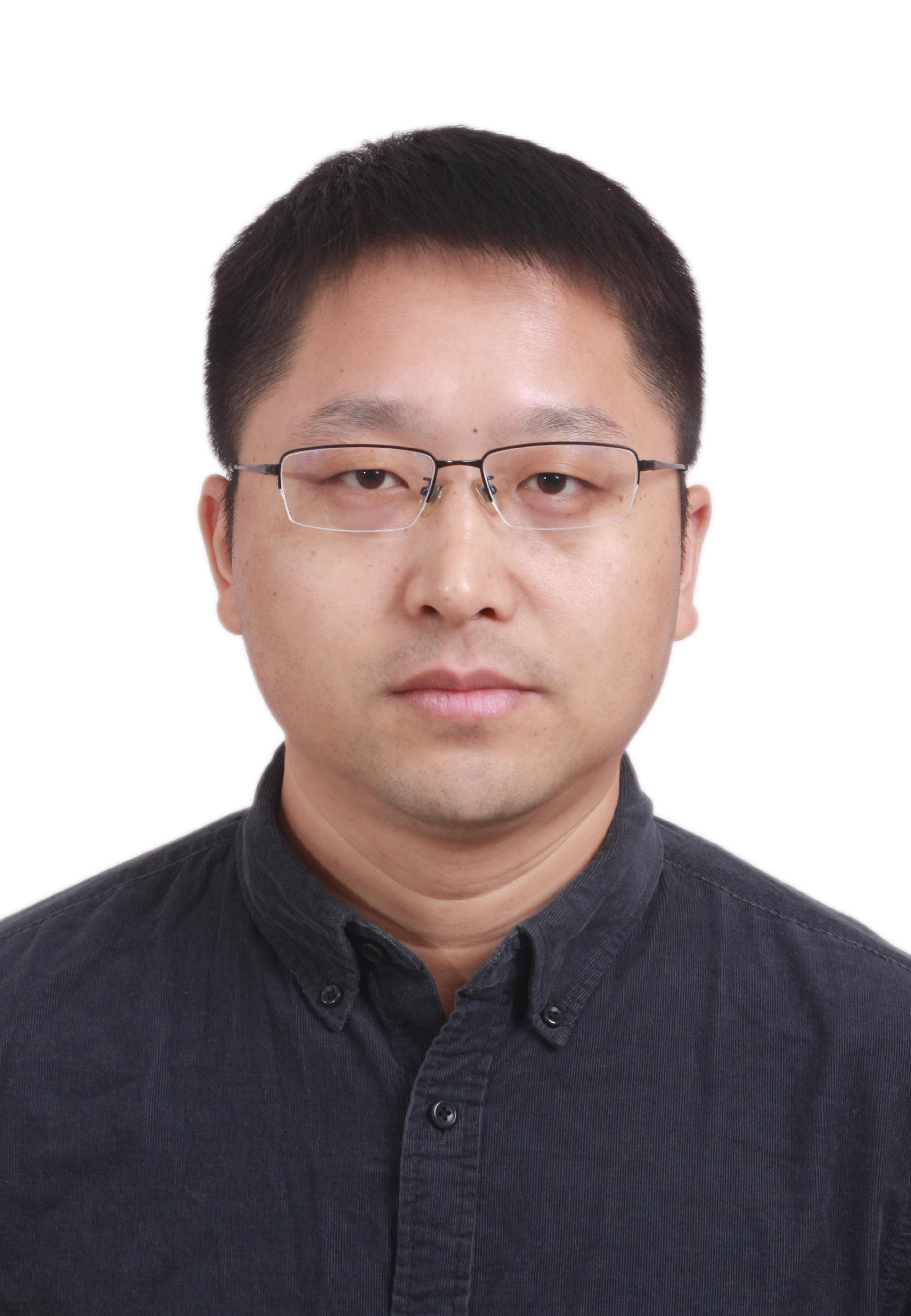}}]{Xianfei Pan} received the Ph.D. degree in control science and engineering from the National University of Defense Technology, Changsha, China, in 2008. Currently, he is a professor of the College of Intelligence Science and Technology, National University of Defense Technology. His current research interests include Inertial navigation system and indoor navigation system.
\end{IEEEbiography}

\end{document}